\def\year{2022}\relax
\documentclass[letterpaper]{article} 
\usepackage{amsthm}
\newtheorem{definition}{Definition}
\usepackage{algorithmic}
\usepackage{textcomp}
\usepackage{xcolor}
\def\BibTeX{{\rm B\kern-.05em{\sc i\kern-.025em b}\kern-.08em
    T\kern-.1667em\lower.7ex\hbox{E}\kern-.125emX}}

\usepackage{subfig} 
\usepackage{mathtools}
\usepackage[utf8]{inputenc}
\usepackage[english]{babel}

\usepackage[export]{adjustbox}
\usepackage{algorithm,algorithmic}
\usepackage{enumitem}
\usepackage{textcomp}
\usepackage{xcolor}
\usepackage{comment}
\usepackage{subfig}
\usepackage{adjustbox}
\usepackage{array}
\usepackage{booktabs}
\usepackage{multirow}
\usepackage{hhline}
\usepackage{mathtools}
\usepackage{dsfont}

\usepackage{times}
\usepackage{soul}
\newtheorem{theorem}{Theorem}
\usepackage[utf8]{inputenc}
\usepackage{booktabs}
\DeclareMathOperator*{\argmin}{arg\,min}

\usepackage{times}
\usepackage{helvet}
\usepackage{courier}
\usepackage{amssymb,amsfonts}
\usepackage{soul}
\usepackage[utf8]{inputenc}
\usepackage{amsmath}
\usepackage{booktabs}
\usepackage[utf8]{inputenc}
\usepackage[english]{babel}
\usepackage{amsmath}
\usepackage{times}
\usepackage{helvet}
\usepackage{courier}

\usepackage{lipsum}
\usepackage[export]{adjustbox}
\usepackage{algorithm,algorithmic}
\usepackage{enumitem}
\usepackage{textcomp}
\usepackage{bbm}
\usepackage[utf8]{inputenc}
\usepackage{xcolor}
\usepackage{comment}

\usepackage[export]{adjustbox}
\usepackage{algorithm,algorithmic}
\usepackage[utf8]{inputenc}
\usepackage{lipsum}

\newcommand\blfootnote[1]{%
  \begingroup
  \renewcommand\thefootnote{}\footnote{#1}%
  \addtocounter{footnote}{-1}%
  \endgroup
}

\usepackage{aaai22}  
\usepackage{times}  
\usepackage{helvet}  
\usepackage{courier}  
\usepackage[hyphens]{url}  
\usepackage{graphicx} 
\usepackage{natbib}  
\usepackage{caption} 
\DeclareCaptionStyle{ruled}{labelfont=normalfont,labelsep=colon,strut=off} 
\usepackage{algorithm}
\usepackage{algorithmic}

\usepackage{newfloat}
\usepackage{listings}
\floatstyle{ruled}
\newfloat{listing}{tb}{lst}{}
\floatname{listing}{Listing}
\title{Fairness without the sensitive attribute via Causal Variational Autoencoder}
\title{Fairness without the sensitive attribute via Causal Variational Autoencoder}
\author {
    Vincent Grari,\textsuperscript{\rm 1,2}
    Sylvain Lamprier, \textsuperscript{\rm 1}
    Marcin Detyniecki \textsuperscript{\rm 2,3}
}
\affiliations {
    \textsuperscript{\rm 1} Sorbonne Universit\'{e}, LIP6/CNRS, Paris, France\\
    \textsuperscript{\rm 2}AXA, Paris, France\\
    \textsuperscript{\rm 3} Polish Academy of Science, IBS PAN, Warsaw, Poland\\
    
vincent.grari@lip6.fr,
sylvain.lamprier@lip6.fr,
marcin.detyniecki@axa.com
}

\usepackage{bibentry}

\begin{document}

\maketitle

\begin{abstract}

In recent years, most fairness strategies in machine learning models focus on mitigating unwanted biases by assuming that the sensitive information is observed. However this is not always possible in practice. Due to privacy purposes and various regulations such as RGPD in EU, many personal sensitive attributes are frequently not collected. We notice a lack of approaches for mitigating bias in such difficult settings, 
in particular for achieving classical fairness objectives such as Demographic Parity and Equalized Odds.
By leveraging recent developments for approximate inference, we propose an approach to fill this gap.
Based on a causal graph, we rely on a new variational auto-encoding based framework named SRCVAE 
to infer a sensitive information proxy, that serve for bias mitigation in an adversarial fairness approach. 
We empirically 
demonstrate significant improvements over 
existing works in the field. We observe 
that the generated proxy's latent space recovers sensitive information and that our approach achieves a higher accuracy while obtaining the same level of fairness on two real datasets, as measured using common fairness definitions.







\end{abstract}

\section{Introduction}


Over the past few years, machine learning algorithms have emerged in many different fields of application.
However, 
there is a growing concern about their potential to reproduce discrimination against a particular group of people based on sensitive characteristics such as religion, race, gender, or other. In particular, algorithms trained on biased data are prone to learn, perpetuate or even reinforce these biases~\cite{NIPS2016_6228}. 
Numerous incidents of this nature have been reported~\cite{angwin2016machine,Lambrecht} in recent years. 
For this reason, there has been a dramatic rise of interest for fair machine learning by the academic community and 
many bias mitigation strategies have been proposed~\cite{zhang2018mitigating,adel2019one,Hardt2016,grari2019fairness,chen2019fairness,zafar2015fairness,celis2019classification,wadsworth2018achieving} during the last decade.
%
Currently, most 
state-of-the-art in fair machine learning algorithms require the knowledge of sensitive information during training. However, in practice, it is unrealistic to assume that sensitive information is available and even collected. In Europe, for example, a car insurance company cannot ask a potential client about his or her origin or religion, as this is strictly regulated. 
The EU introduced the General Data Protection Regulation (GDPR) in May 2018. This legislation has  represented one of the most important changes in the regulation of data privacy from more than 20 years. It strictly regulates the collection and use of sensitive personal data. With the aim of obtaining non-discriminatory algorithms,
Article 9(1) rules that in general: "Processing of personal data revealing racial or ethnic origin, political opinions, religious or philosophical beliefs, or trade union membership, and the processing of genetic data, biometric data for the purpose of uniquely identifying a natural person, data concerning health or data concerning a natural person's sex life or sexual orientation shall be prohibited."~\cite{citeulike:14071352}. 
Ignoring sensitive attributes as input of predictive machines is known as "fairness through  unawareness"~\cite{Pedreshi2008}.
However, this is clearly not sufficient, since  complex correlations in the data may provide unexpected links to sensitive information. In fact, presumably non-sensitive attributes can serve as substitutes or proxies for protected attributes. For instance, the color and the model of a car combined with the driver's occupation can lead to unwanted gender bias in the prediction of car insurance price. To cope with it, a large trend of machine learning methods has recently emerged to mitigate biases in output predictions, based on some fairness criteria w.r.t. the sensitive attributes \cite{zhang2018mitigating,adel2019one,Hardt2016,grari2019fairness}. 
However, all these methods use the sensitive during training which is not always possible.

Recently, some works addressed this complex objective to obtain a fair predictor model without the availability of the sensitive. Most of these works leverage the use of 
external data or prior knowledge on correlations~\cite{zhao2021you,madras2018learning,schumann2019transfer,gupta2018proxy}. 
Some other methods avoid the need of such additional data, but are only based on some local smoothness of the feature space, rather than explicitly focusing on targeted subgroups to be protected~\cite{hashimoto2018fairness,lahoti2020fairness}. 
The need of new approaches for algorithmic fairness that break away from the prevailing assumption of observing sensitive characteristics to be fair has also been recently highlighted in \cite{tomasev2021fairness}. 

To fill the gap, we propose a novel approach that relies on bayesian variational autoencoders (VaEs) to infer the sensitive information given a causal graph provided by expert knowledge of the data, 
and then uses the inferred information as proxy for mitigating biases in a adversarial fairness training setting. More specifically, the latter mitigation is performed during predictor training, by considering a fairness loss based on an estimation of the correlation between predictor outputs and the inferred sensitive proxies. 
We strive in this paper for showing the empirical interest of this type of method. 

\noindent





\section{Problem Statement}
\label{sec:Problem_statement}


Throughout this document, we consider a supervised machine learning algorithm for classification problems. The training data consists of $n$ examples ${(x_{i},s_{i},y_{i})}_{i=1}^{n}$, where $x_{i} \in \mathbb{R}^{p}$ is the feature vector with $p$ predictors of the $i$-th example, $s_i$ is its binary discrete sensitive attribute and $y_{i}$ its binary discrete outcome true value. 


\subsubsection{Demographic Parity}
\label{sec:demographic_parity}
A classifier is considered fair if the prediction $\widehat{Y}$ from features $X$ is independent from the protected attribute $S$~\cite{Dwork2011}. The underlying idea is that each demographic group has the same chance for a positive outcome.

\begin{definition}
$P(\widehat{Y}=1|S=0)=P(\widehat{Y}=1|S=1)$
\end{definition}



There are multiple ways to assess this objective.
The p-rule assessment ensures the ratio of the positive rate for the unprivileged group is no less than  a fixed threshold $\frac{p}{100}$.
The classifier is considered as totally fair when this ratio satisfies a 100\%-rule. Conversely, a 0\%-rule indicates a completely unfair model.
{\small
\begin{equation}
\emph{P-rule}(\widehat{Y},S)=
\min(\frac{P(\widehat{Y}=1|S=1)}{P(\widehat{Y}=1|S=0)},\frac{P(\widehat{Y}=1|S=0)}{P(\widehat{Y}=1|S=1)})
\nonumber
\end{equation}
}
\subsubsection{Equalized Odds}
\label{sec:equalized_odds}
An algorithm is considered fair if across both demographics $S=0$ and $S=1$,   
the predictor $\widehat{Y}$ has equal \textit{false} positive rates, and 
\textit{false} negative rates~\cite{Hardt2016}. This constraint enforces that accuracy is equally high in all demographics since the rate of positive and negative classification is equal across the groups. The notion of fairness here is that chances of being correctly or incorrectly classified positive should be equal for every group.
\begin{definition}
$P(\widehat{Y}=1|S=0,Y=y)=P(\widehat{Y}=1|S=1,Y=y), \forall y\in\{0,1\}$
\end{definition}




A metric to assess this objective is to measure the disparate mistreatment (DM)~\cite{zafar2015fairness}. It computes the absolute difference between the false positive rate (FPR) and the false negative rate (FNR) for both demographics.
{\small
\begin{eqnarray}
\Delta_{FPR}:
|P(\widehat{Y}=1|Y=0,S=1)-P(\widehat{Y}=1|Y=0,S=0)|
\nonumber
\\
\Delta_{FNR}:
|P(\widehat{Y}=0|Y=1,S=1)-P(\widehat{Y}=0|Y=1,S=0)|
\nonumber
\end{eqnarray}
}
The closer the values of $\Delta_{FPR}$ and $\Delta_{FNR}$ to 0, the lower the degree of disparate mistreatment of the classifier.
 
In this paper we deal with those aforementioned fairness metrics for problems where the sensitive is hidden at train time.

\section{Related work}
\label{sec:rw}

From the state-of-the-art literature, one possible way to overcome the unavailability 
of sensitive attributes during training 
is to use transfer learning 
methods, 
from external sources of data where the sensitive group labels are known. For example, \cite{madras2018learning} proposed to learn fair representations via adversarial learning on a specific downstream task and transfer it to the targeted one. 
\cite{schumann2019transfer} and \cite{coston2019fair} focus on domain adaptation. 
\cite{mohri2019agnostic} consider an agnostic federated learning where a centralized model is optimized for any possible target distribution
formed by a mixture of different clients distributions. However, the actual desired bias mitigation is highly dependent on the distribution of the external source of data. 


Another trend of works requires prior knowledge on sensitive correlations. With prior assumptions, 
\cite{gupta2018proxy} and \cite{zhao2021you} 
mitigate the 
dependence 
of the predictions on the  available features 
that are known to be likely correlated with the sensitive. However, 
such strongly correlated features do not always exist in the data. 

Finally, some approaches address this objective without any 
prior knowledge on the sensitive. First, some approaches aim at improving the accuracy for the worst-case protected group for achieving the Rawlsian Max-Min fairness objective. 
This implies techniques from distributionally robust optimization \cite{hashimoto2018fairness} or adversarial learning \cite{lahoti2020fairness}.
Other approaches such as \cite{yan2020fair} act on the input of the data via a cluster-based balancing strategy. These methods have the advantage to require no additional sensitive data or knowledge, but are often ineffective for traditional group fairness definitions such as \emph{demographic parity} and \emph{equalized odds}. Their blind way of mitigation usually implies a strong degradation of the predictor accuracy, by acting on non-sensitive information.

Our approach is inherently different from the
aforementioned approaches. Based on light prior knowledge on some causal relationships in the data, we rely on the bayesian inference  
of latent sensitive proxies, whose dependencies with model outputs can then be easily mitigated in  a second training step. 

\section{Methodology}
\label{sec:methodology}

We describe in this section our methodology to provide a fair machine learning model 
for training data without sensitive demographics. For this purpose, we first assume some specific causal graph which underlies training data. 
This causal graph allows us to infer, through Bayesian Inference, a latent representation containing as much information as possible about the sensitive feature. 
Finally, this section presents 
our methodology to mitigate fairness biases, 
while maintaining as 
prediction accuracy as possible for the model. 


\subsection{Causal Structure of the Sensitive Retrieval Causal Variational Autoencoder: SRCVAE}
\label{causal_stucture}

\begin{figure}
  \centering
  \includegraphics[scale=0.39]{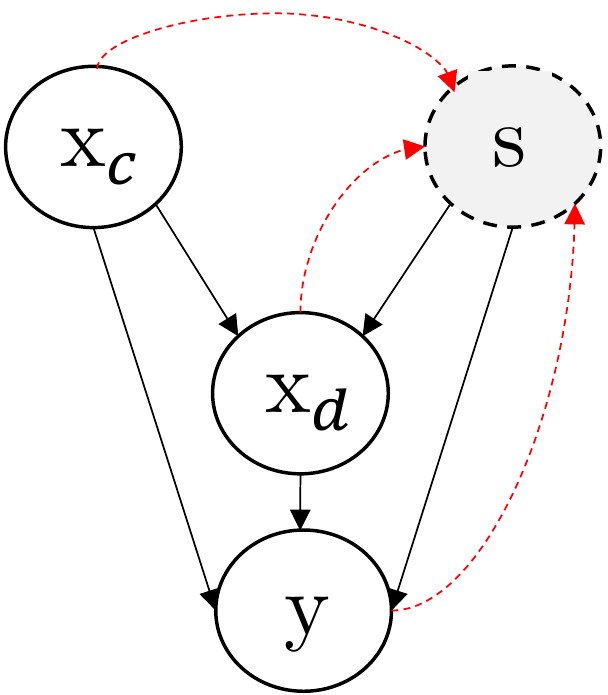}
  \includegraphics[scale=0.39]{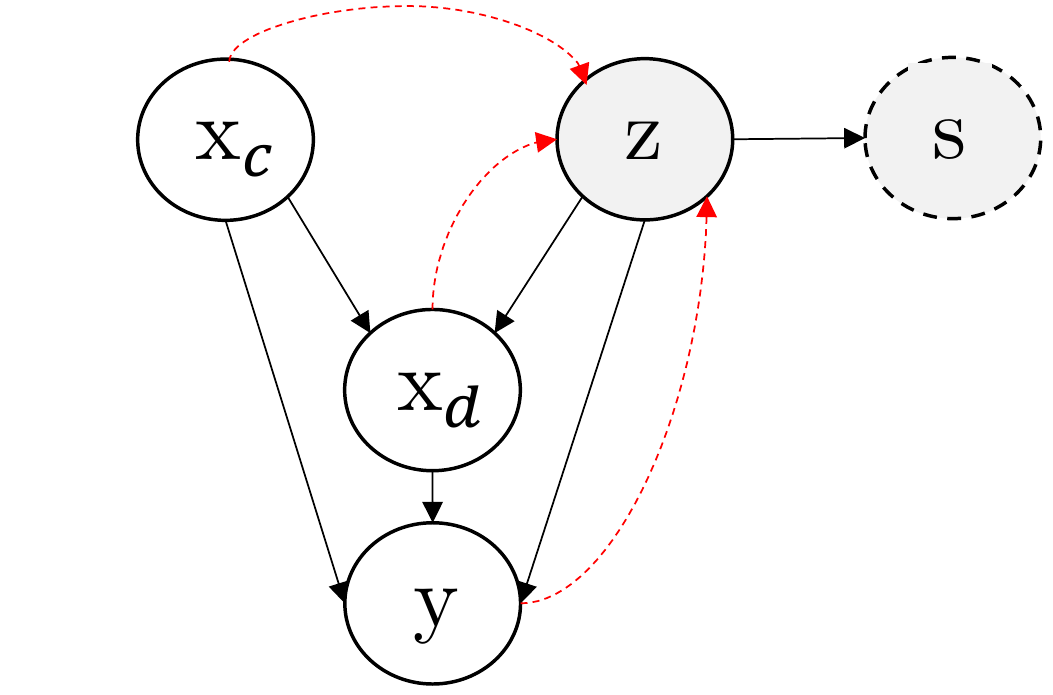}
  \caption{Causal graphs of SRCVAE: Left graph represents prior expert knowledge  of the data, where $x$ is mapped into two components $x_c$ and $x_d$. Right graph denotes the graph considered in our approach, with a multivariate confounder $z$ infered to be used as a proxy of the sensitive $s$. 
  Solid lines denote the generative process, arrows with red dashed lines denote inference, grey circles denote the missing attributes.}
  
  
  \label{fig:causal_graph}
\end{figure}

\begin{figure}[t]
    \centering
    \includegraphics[scale=0.36]{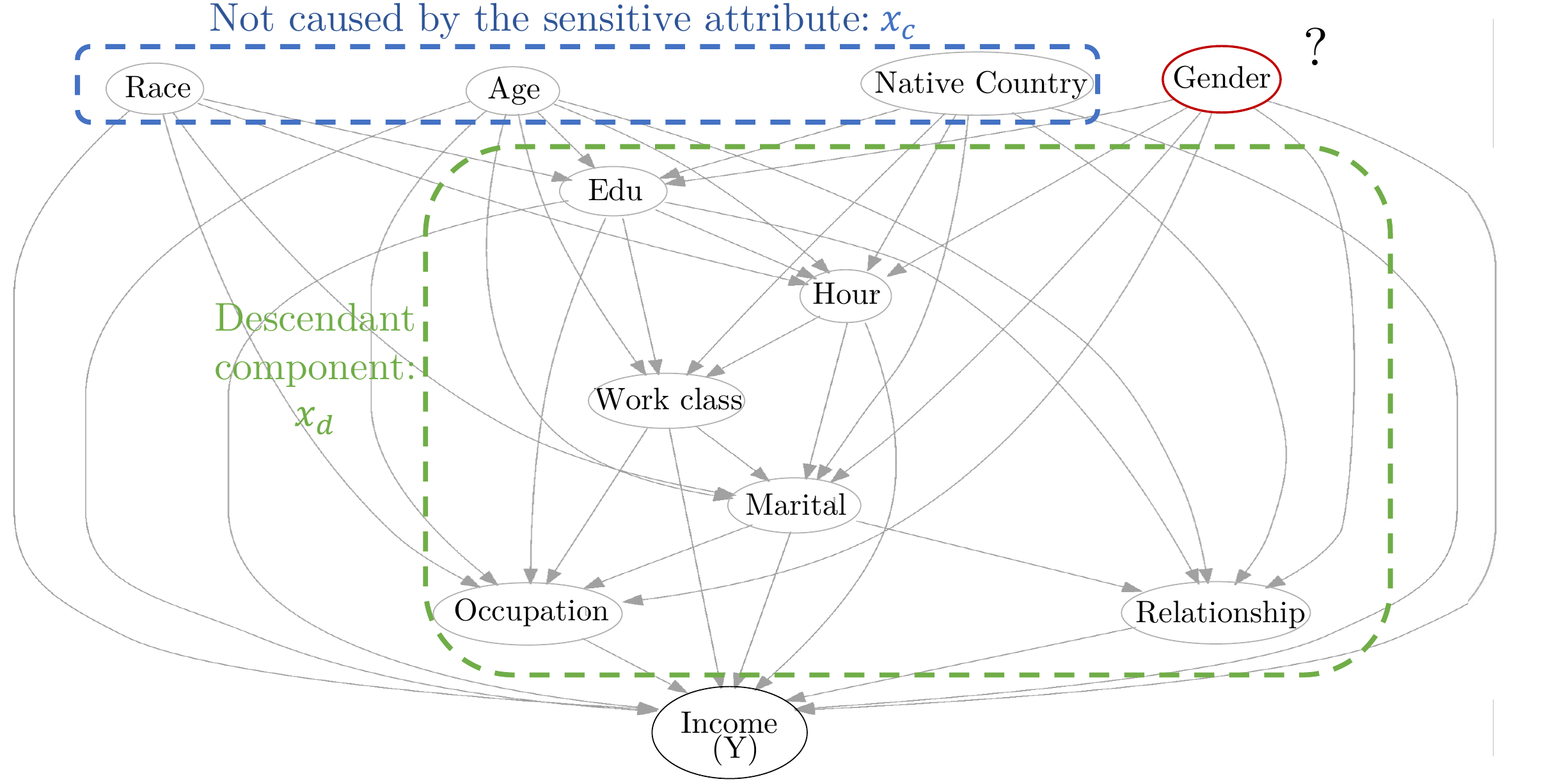}
    \captionof{figure}{Causal Graph - Adult UCI}
    \label{fig:adult_causalgraph}
\end{figure}

Our work relies on the assumption of underlying causal graphs given in 
Figure \ref{fig:causal_graph}, which fit with many real world settings (slightly different graphs are studied in appendix). 
In the leftmost graph, parents of the output $y$ are split in three  components $x_c$, $x_{d}$ and $s$, where $x_c$ contains only variables not caused by the sensitive attribute 
$s$.
The variables subset $x_d$ 
is both caused by the sensitive information $s$ and $x_c$. 
For instance, in Figure~\ref{fig:adult_causalgraph}, if we regard the assumed causal graph of the Adult UCI dataset  with Gender as the sensitive attribute $s$ and Income as the expected output $y$, $x_c$ is the set 
of variables \emph{Race}, \emph{Age} and \emph{Native\_Country} which do not depend on the sensitive,  and $x_{d}$ corresponds to all remaining variables that are generated from $x_c$ and $s$ (i.e.,   
$x_{d} = \{Education, Work\_Class, ...\}$). 

Assuming the observation of all variables except $S$, our purpose in the following is to recover all the hidden information not caused by the set $x_c$ but responsible of $x_d$ and $y$. 
In a real world scenario, it is noteworthy that the accuracy with which one can recover the real sensitive $s$ 
 depends on 
 the right representation of the complementary set $x_c$. 
If the set $x_c$ is under-represented, the  
reconstruction of $s$ (whose methodology is given in the next section) may contain additional   
information. 
Assuming that the graph from Figure~\ref{fig:adult_causalgraph} is the exact causal graph that underlies the Adult UCI, and imagine a setting where for instance the variable Race is hidden, 
this variable is likely to leak in the sensitive reconstruction. 
%
We argue that reconstructing a binary sensitive feature with other leakage information strongly degrades the inferred sensitive proxy. This is what motivated us to rather consider the rightmost graph from \ref{fig:causal_graph} approach, that considers 
a multivariate continuous intermediate confounder $z$ which both causes the sensitive $s$ and the observed variables in $x_d$ and $y$. As we observe in the experiments section, such a multivariate proxy also allows better generalization abilities for mitigated prediction.  

\subsection{Sensitive Reconstruction} 
\label{Elbo_VAE}

We describe in this section 
the first step of our SRCVAE framework for generating a latent representation $Z$ which contains as much as possible information about the real sensitive feature $S$. As discussed above, our strategy is to use Bayesian inference approximation, according to the 
pre-defined causal graph in Figure \ref{fig:causal_graph}. 

A simple methodology for performing inference 
could be to suppose with strong hypothesis a non deterministic structural model with some specific distribution for all the causal links and finally estimate the latent space by probabilistic programming language such as Stan \cite{kusner2017counterfactual,team2016rstan}.  
Leveraging recent developments for approximate inference with deep learning, many different works \cite{louizos2017causal,pfohl2019counterfactual,grari2020adversarial} 
proposed to use Variational Autoencoding~\cite{kingma2013auto} methods (VAE) for modeling exogenous variables in causal graphs. It has shown successful empirical results and in particular for the counterfactual fairness sub-field.
The counterfactual objective is different from our objective since the inference is performed for generating some exogenous variables 
independently from the sensitive. It aims at building prediction models which ensure fairness at the most individual level, 
by 
requiring the notion of the $do$ operator \cite{pearl2009causal} with
the intervention of the sensitive in causality ($do(s=1)$ or $do(s=0)$). Notice here, we don't require this notion since we are only interested to capture a stochastic exogenous variables by generating a sensitive proxy distribution for each individual.



Following the rightmost causal graph from Figure~\ref{fig:causal_graph}, 
the decoder distribution  $p_{\theta}(x_{c}, x_{d}, y|z)$ can be factorized as below:
\begin{align}
p_{\theta}(x_{c}, x_{d}, y|z) = p(x_{c})p_{\theta}(x_{d}|x_{c},z)p_{\theta}(y|x_{c},x_{d},z) \nonumber
\end{align}


Given an approximate posterior $q_{\phi}(z|x_{a},x_{d},y)$, we
obtain the following variational lower bound: 
\begin{align}
log(p_{\theta}(x_{c}, x_{d}, y)) \geq & \mathbb{E}_{\substack{(x_{c},x_{d},y)\sim\mathcal{D}, \\ z \sim q_{\phi}(z|x_{c},x_{d},y)}}[\log p_{\theta}(x_{d},y|x_{c},z)
\nonumber \\
& 
- D_{KL}(q_{\phi}(z|x_{c},x_{d},y)||p(z))\big] \nonumber \\
&=: -\mathcal{L}_{ELBO}
\label{equ:elbo}
\end{align}
where $D_{KL}$ denotes the Kullback-Leibler divergence of the posterior $q_{\phi}(z|x_{c},x_{d},y)$ from a prior $p(z)$, typically a standard Gaussian distribution ${\cal N}(0,I)$. The posterior  $q_{\phi}(z|x_{c},x_{d},y)$ is represented by a deep neural network with parameters $\phi$, which typically outputs the mean $\mu_\phi$ and the variance $\sigma_\phi$ of a diagonal Gaussian distribution ${\cal N}(\mu_\phi,\sigma_\phi I)$.

The likelihood term which factorizes as  
$p_{\theta}(x_{d},y|x_{c},z)=  p_{\theta}(x_{d}|x_{c},z)p_{\theta}(y|x_{c},x_d,z)$, is defined as neural networks 
with parameters $\theta$. 
Since attracted by a standard prior, the posterior is supposed to remove probability mass for any features of  
$z$ that are not involved in the reconstruction of $x_d$ and $y$. Since $x_{c}$ is given together with $z$ as input of the likelihoods, all the information from $x_{c}$ should be removed from the posterior  distribution of $z$.
We employ in this paper a variant of the ELBO optimization as done in  \cite{pfohl2019counterfactual}, where the $D_{KL}(q_{\phi}(z|x_{c},x_d,y)||p(z))$ term is replaced by a MMD term $\mathcal{L}_{MMD}(q_{\phi}(z)||p(z))$ between the aggregated posterior $q_{\phi}(z)$ and the prior. This has been shown more powerful than the classical $D_{KL}$  
for ELBO optimization in \cite{zhao2017infovae}, as the latter can reveal as 
too restrictive (uninformative latent code problem) \cite{chen2016variational,bowman2015generating,sonderby2016ladder} and can also tend to overfit the data (Variance Over-estimation in Feature Space).


This inference must however ensure that no dependence is created between $x_{c}$ and $z$ (no arrow from $x_{c}$ to $z$ in the rightmost graph~\ref{fig:causal_graph}, 
to prevent the generation of proper sensitive proxy which is not linked to the complementary. 
However, 
by optimizing this ELBO equation
, some dependence can still be observed empirically between $x_{c}$ and $z$ as we show through our experimental results part. 
Some information from $x_{c}$ leaks in the inferred $z$. In order to ensure some minimum independence level we add a penalisation term in this loss function. Leveraging recent research for mitigating the dependence between continuous 
variables, we extend the main idea of \cite{grari2020learning,grari2019fairness} by adapting this penalization in the variational autoencoder case. Following the idea of~\cite{grari2020learning}, we consider the HGR coefficient as defined in definition \ref{hgr} to ensure the independence level. 


\begin{definition}
For two jointly distributed random variables $U \in \mathcal{U}$ and $V \in \mathcal{V}$
, the Hirschfeld-Gebelein-R\'enyi maximal correlation is
defined as:
\begin{align}
HGR(U, V) &= \sup_{\substack{ f:\mathcal{U}\rightarrow \mathbb{R},g:\mathcal{V}\rightarrow \mathbb{R}}} \rho(f(U), g(V)) \\
&= \sup_{\substack{ f:\mathcal{U}\rightarrow \mathbb{R},g:\mathcal{V}\rightarrow \mathbb{R}\\
           E(f(U))=E(g(V))=0 \\   E(f^2(U))=E(g^2(V))=1}} E(f(U)g(V))
\label{hgr}
\end{align}
where $\rho$ is the Pearson linear correlation coefficient 
with some measurable functions $f$ and $g$ with positive and finite variance. 
\end{definition}
The HGR coefficient is equal to $0$ if the two random variables are independent, and is equal to 1 if they are strictly dependent. 
The HGR estimation is performed 
via two inter-connected neural networks by approximating the optimal transformation functions $f$ and $g$ from (\ref{hgr}) as \cite{grari2019fairness,grari2020learning}.

In the following, we denote as $\mathop{\widehat{HGR}_{w_f,w_g}}\limits_{U\sim\mathcal{D_U},V\sim\mathcal{D_V}}(U, V)$ the neural estimation of HGR computed as:
\begin{align}
\nonumber
\mathop{\widehat{HGR}_{w_f,w_g}}_{U\sim\mathcal{D_U},V\sim\mathcal{D_V}}(U, V) &=\mathop{max}_{w_f,w_g}\mathbb{E}_{U\sim\mathcal{D_U},V\sim\mathcal{D_V}}(\widehat{f}_{w_{f}}(U)\widehat{g}_{w_{g}}(V))
\label{hgr}
\end{align}
where $\mathcal{D_U}$ and $\mathcal{D_V}$ represent the respective distributions of $U$ and $V$. 
The neural network $f$ with parameters $w_f$ takes as input the variable $U$ and the neural network $g$ with parameters $w_g$ takes as input $V$. 
At each iteration, the estimation algorithm first standardizes the output scores of networks $f_{\omega_f}$ and $g_{\omega_g}$ to ensure 0 mean and a variance of 1 on the batch. Then it computes the objective function to maximize the HGR estimated score. 


Finally, the inference of our SRCVAE approach is optimized by a mini-max game as follows: 

\begin{eqnarray}
\begin{aligned}
\nonumber
    \argmin_{\theta,\phi}\max_{w_{f},w_{g}}&-\mathbb{E}_{\substack{(x_{c},x_{d},y)\sim\mathcal{D}, \\ z \sim q_{\phi}(z|x_{c},x_{d},y)}}[\log p_{\theta}(x_{d},y|x_{c},z)
    \\
    &+\lambda_{mmd}\mathcal{L}_{MMD}(q_{\phi}(z)||p(z))]  \\
    &+
    \lambda_{inf}\mathop{\widehat{HGR}_{w_f,w_g}}_{\substack{(x_{c},x_{d},y)\sim\mathcal{D},\\z \sim q_{\phi}(z|x_{c},x_{d},y)}}(x_c, z)
    \label{FinalInference}
\end{aligned}
\end{eqnarray}

\begin{figure}
\centering
  \subfloat[Max Phase: HGR Estimation]{
  \includegraphics[scale=0.47]{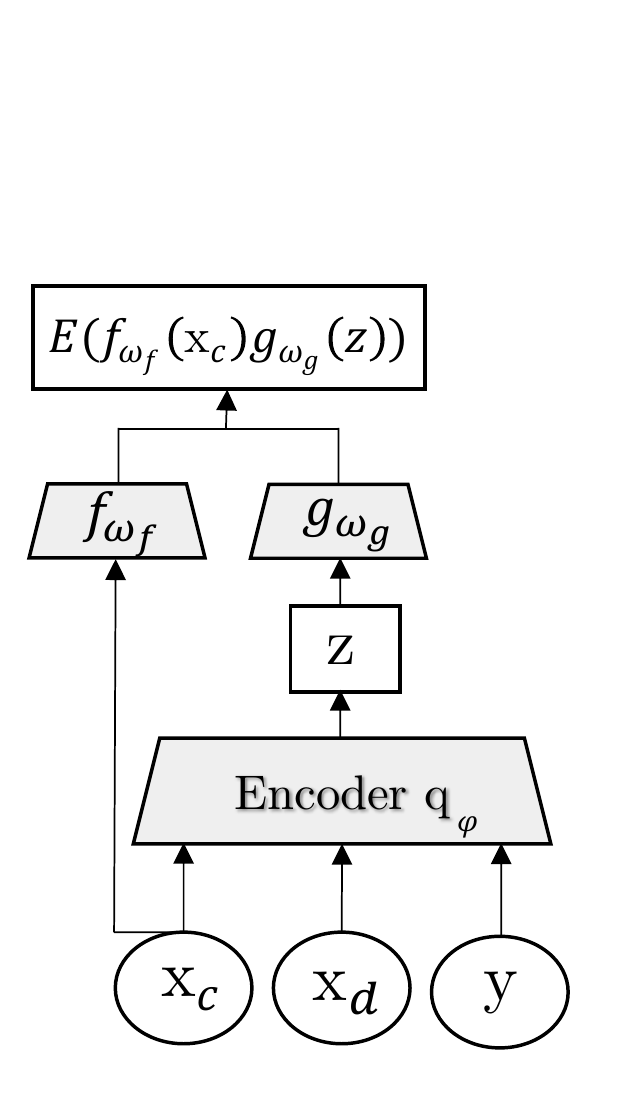}}
  \subfloat[Min Phase: Reconstruction  ]{\includegraphics[scale=0.4]{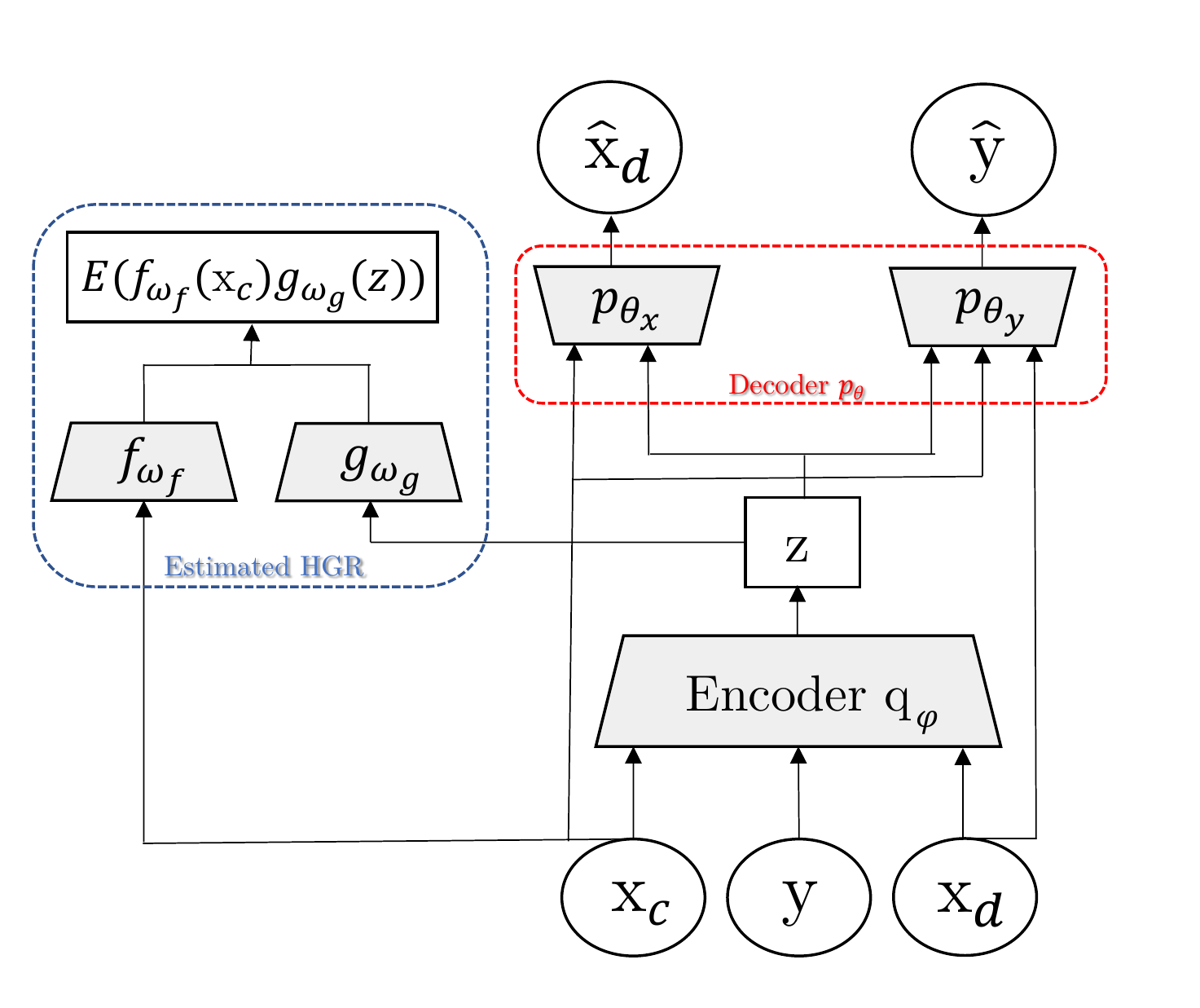}}
  \caption{Neural network architecture of SRCVAE in max phase for the HGR estimation between $a$ and $z$ via gradient ascent (a) and Variational autoencoder structure of SRCVAE in min phase for the reconstruction of $x_d$ and $y$ via the generation of the latent space $z$ where the mitigation dependence with $a$ is performed from the estimated HGR.}
  \label{fig:SRCVAE}
\end{figure}

where $\lambda_{mmd}$, $\lambda_{inf}$ are scalar hyperparameters. 
The additional MMD objective  
can be interpreted as minimizing the distance between all moments of each aggregated latent code distribution and the prior distribution. 
Note that 
the use of $y$ as input for our generic inference scheme $q(z|x_{c},x_d,y)$ is allowed since $z$ is only used during training for learning a fair predictive model in the next section 
and is not used at deployment time. 

In figure~\ref{fig:SRCVAE}, we represent the min-max structure of SRCVAE. The left architecture represents the max phase where the HGR between 
$z$ and $x_{c}$ is estimated by gradient ascent with multiple iterations. The right graph represents the min phase where the reconstruction of $x_d$ and $y$ is performed by the decoder $p_\theta$ (red frame) via the generated latent space $z$ from the decoder $q_\phi$. The 
adversarial HGR component (blue frame) ensures the independence level between the generated latent space $z$ and $x_{c}$. The adversarial $f$ takes as input the set $x_{c}$ and the 
adversarial $g$ takes the continuous latent space $z$.
In that way, we 
capture for each gradient iteration the estimated HGR between the set $x_c$ and the generated proxy latent space $z$. 
At the end of each iteration, the algorithm updates the parameters of the decoder parameters $\theta$ as well as the encoder parameters $\phi$ by one step of gradient descent. Concerning the HGR adversary, the backpropagation of the parameters $\omega_f$ and $\omega_g$ is performed by multiple steps of gradient ascent. This allows us to optimize a more accurate estimation of the HGR at each step, leading to a greatly more stable learning process. The hyperparameter $\lambda_{inf}$ controls the impact of the dependence loss in the optimization.

\subsection{Mitigate the unwanted biases}
\label{mitigation}

Once a sensitive proxy is available from the inference  method of the previous section, 
the goal is now to use it 
for training a fair predictive function $h_\theta$. 
Since $z$ contains some continuous multidimensional information, we adopt an HGR-based approach inspired from \cite{grari2019fairness,grari2020learning} which have shown superior performance in this context. We also observe this claim empirically in our context, results are shown in appendix. 



 
Depending on the fairness objectives, we propose to mitigate the unwanted bias via an adversarial penalization during the training phase.
\subsubsection{Demographic Parity}

We propose to find a mapping $h_\theta(x_{c},x_{d})$ which both minimizes the  deviation with the expected target $y$ and does not imply too much dependency with the representation $z$. This information proxy $z$ is generated as mentioned above from the posterior distribution $q_{\phi}(z|x_{c},x_{d},y)$. This enables to generate a sensitive distribution proxy for each individual where the dependence with the output prediction is assessed and mitigated all along of the training phase. 
In this paper, we extend the idea of \cite{grari2019fairness} by proposing a novel neural HGR-based cost for fairness without demographics via inference generation. We propose the 
optimization problem as follows:

\begingroup
\begin{eqnarray}
\begin{aligned}
\nonumber
    \arg\min_{\theta}\max_{\psi_{f},\psi_{g}}&\mathcal{L}(h_\theta(x_{c},x_{d}),y) \\ &+
    \lambda_{DP}\mathop{\widehat{HGR}_{\psi_f,\psi_g}}_{\substack{(x_{c},x_{d},y)\sim\mathcal{D},\\z \sim q_{\phi}(z|x_{c},x_{d},y)}}(h_\theta(x_{c},x_{d}), z)
    \label{genericfunctionDP}
\end{aligned}
\end{eqnarray}

\endgroup
where  $\mathcal{L}$ is the predictor loss function (the log-loss function in our experiments) between the output $h_\theta(x_{c}, x_{d}) \in \mathbb{R}$ and the corresponding target $y$, with $h_\theta$ a neural network with parameters $\theta$ which takes as input the 
set $x_{c}$ and the descendant attribute $x_{d}$. 
The hyperparameter $\lambda_{DP}$ controls the impact of  
the dependence between the output prediction $h_\theta(x_{d},x_{c})=p_{\theta}(y=1|x_{d},x_{c})$ and the sensitive proxy $z$. 
For each observation $(x_{c_{i}},x_{d_{i}},y_{i})$, we generate $K$ (200 in our experiment) different latent variables $z^{k}_{i}$ ($k$-ith generation) from the causal model. 
As in the inference phase, the backpropagation of the HGR adversary with parameters $\psi_f$ and $\psi_g$ is performed by multiple steps of gradient ascent. This allows us to optimize an accurate estimation of the HGR at each step, leading to a greatly more stable predictive learning process.

\subsubsection{Practice in real-world}
\label{practise_in_real_world}
As mentioned in the first subsection
, the assumed causal graph~\ref{fig:causal_graph} requires the right representation of the complementary set $x_{c}$. 
If the set $x_{c}$ is under-represented, 
some specific hidden attributes 
can be integrated with the sensitive in the inferred sensitive latent space $z$.
The following Theorem~\ref{theoremHGR} allows us to ensure that mitigating the 
HGR between $z$ and $\widehat{y}$ implies some upperbound for  
the targeted objective.

\begin{theorem}
For two nonempty index sets $S$ and $Z$ such that $S \subset Z$ and $\hat{Y}$ the output prediction of a predictor model, we have:\\ 
\begin{align}
HGR(\hat{Y},Z) \ge HGR(\hat{Y},S)
\end{align}
\label{theoremHGR}
\end{theorem}
\proof{in appendix}

\subsubsection{Equalized odds}

We extend the demographic parity optimization 
to the equalized odds task. 
The objective is to find a mapping $h_\theta(x_{c},x_{d})$ which both minimizes the deviation with the expected target $y$ and does not imply too much dependency with the representation $z$ conditioned on the actual outcome $y$. 
For the decomposition of disparate mistreatment, the mitigation shall be divided into the two different values of $Y$.
By identifying and mitigating the specific non linear dependence for these two subgroups, it enforces the two objectives of having the same false positive rate and the same false negative rate for each demographic.  

The mitigation of this optimization problem is as follows:

\begingroup
\small 
\begin{eqnarray}
\begin{aligned}
\nonumber
    \arg\min_{\theta}\max_{{\psi_{f_{0}},\psi_{g_{0}},\psi_{f_{1}},\psi_{g_{1}}}}&\mathcal{L}(h_\theta(x_{c},x_{d}),y)
    \\ &+
    \lambda_{0}\mathop{\widehat{HGR}_{\psi_{f_{0}},\psi_{g_{0}}}}_{\substack{(x_{c},x_{d},y)\sim\mathcal{D}_0,\\z \sim q_{\phi}(z|x_{c},x_{d},y)}}(h_\theta(x_{c},x_{d}), z)\\
    &+
    \lambda_{1}\mathop{\widehat{HGR}_{\psi_{f_{1}},\psi_{g_{1}}}}_{\substack{(x_{c},x_{d},y)\sim\mathcal{D}_1,\\z \sim q_{\phi}(z|x_{c},x_{d},y)}}(h_\theta(x_{c},x_{d}), z)
    \label{genericfunctionEO}
\end{aligned}
\end{eqnarray}
\endgroup

Where $\mathcal{D}_{0}$ corresponds to the observations set $(x_{c},x_{d},y)$ where ${y=0}$ and $\mathcal{D}_{1}$ to observations where ${y=1}$. The hyperparameters $\lambda_{0}$ and $\lambda_{1}$ control the impact of the dependence loss for the false positive and the false negative objective respectively. 
The first penalisation (controlled by $\lambda_0$) enforces the independence between the output prediction $h_\theta(x_{d},x_{c})=p_{\theta}(y=1|x_{d},x_{c})$ and the sensitive proxy $z$ only for the cases where $y=0$. It enforces 
the mitigation of the difference of false positive rates between demographics, since at optimum for $\theta^*$ with no trade-off (i.e., with infinite $\lambda_0$) and $(x_{c},x_{d},y)\sim\mathcal{D}_{0}$,  $HGR(h_{\theta^{*}}(x_{d},x_{c}),z)=0$  and implies theoretically: $h_{\theta^{*}}(x_{d},x_{c}) \perp z| y=0$. 
For the second, it enforces the mitigation of the difference of the true positive rate since the dependence loss is performed between the output prediction $h_\theta(x_{d},x_{c})$ and the sensitive only for cases where $y=1$. In consequence, $\Delta_{FNR}$ is mitigated (since FNR=1-TPR). Note that the use of $y$ as input for our generic mitigation to create a fair \emph{equalized odds} classifier predictor is allowed since this mitigation is only used during training and not at deployment time.
In our experiment 


\section{Experimental results}
\label{sec:experimentalresults}

For our experiments, we empirically evaluate the performance of our contribution 
 on real-world data sets where the sensitive $S$ is available. 
This allows to assess the 
fairness 
of the output prediction, obtained without the use of the sensitive,  w.r.t. this ground truth. 
To do so, we use the popular Adult UCI and Default datasets (descriptions in Appendix) often used in fair classification. 

\begin{figure}
\subfloat[
\small{Without the dependence penalization term: $\lambda_{inf}$=$0.0$ ; $HGR(x_c,z)$=$81.7\%$}]{\includegraphics[scale=0.175]{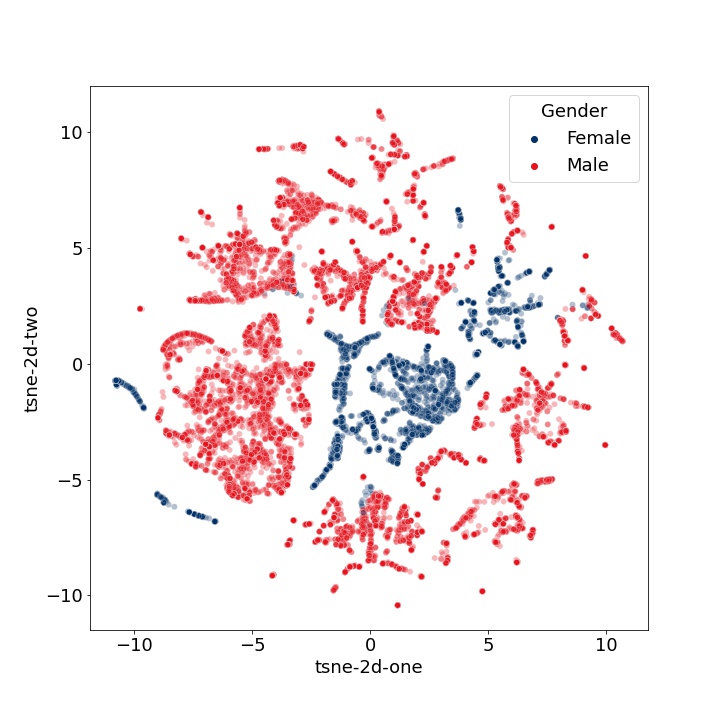}} 
\subfloat[With a dependence term  \hfill  $\lambda_{inf}$=$0.2$ ; $HGR(x_c,z)$=$22.6\%$]{\includegraphics[scale=0.175]{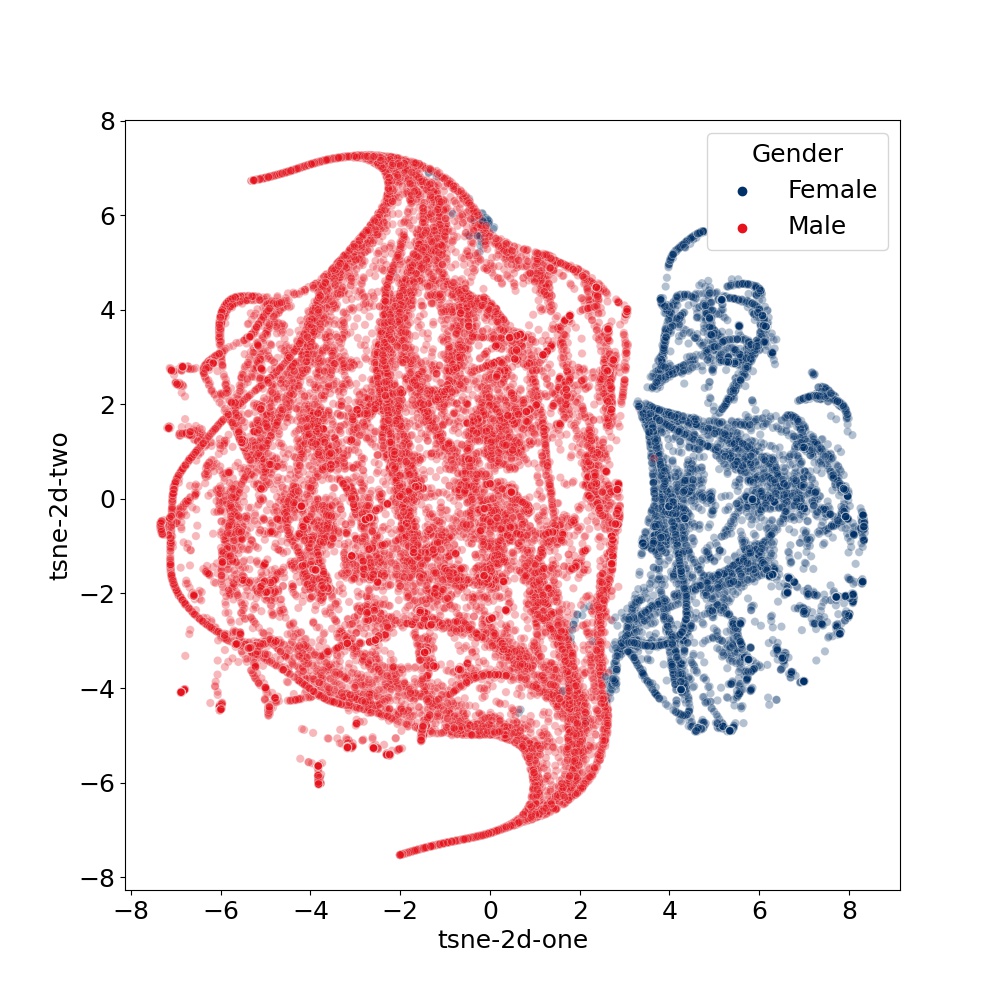}}
\caption{Inference phase for Adult UCI: t-SNE of the sensitive latent reconstruction $Z$. Blue points are males ($S=1$) and red ones are females ($S=0$). 
Increasing the hyperparameter $\lambda_{inf}$ in the inference phase allows to mitigate the dependence between the generated proxy $z$ and the set $x_{c}$ (HGR estimation). We observe a better separation
between the male and female data points which indicates proper sensitive proxy. 
}
\label{fig:tsne}
\end{figure}

For understanding the interest of mitigating the dependence between the latent space $z$ and the complementary set $x_{c}$ during the inference phase, we plot in Figure~\ref{fig:tsne} the t-SNE of $z$ of two different inference models for Adult UCI dataset. 
As a baseline, we consider a version of our model trained without the penalization term ($\lambda_{inf}=0.00$). This is compared to a version trained with  a penalization term equal to $0.20$. 
As expected, training the inference model without the penalization term, 
results in a poor reconstruction proxy $z$, where the dependence with $x_{c}$ is observed. We can observe that the separation is not significant between the data points of men (blue points) and women (red points). 
We also observe that increasing this hyper-parameter allows to decrease the $HGR$ estimation from $81.7\%$ to $22.6\%$ and greatly increase the separation between male and female data points. 

\begin{figure*}[t]
   \centering
   \includegraphics[scale=0.289,valign=t]{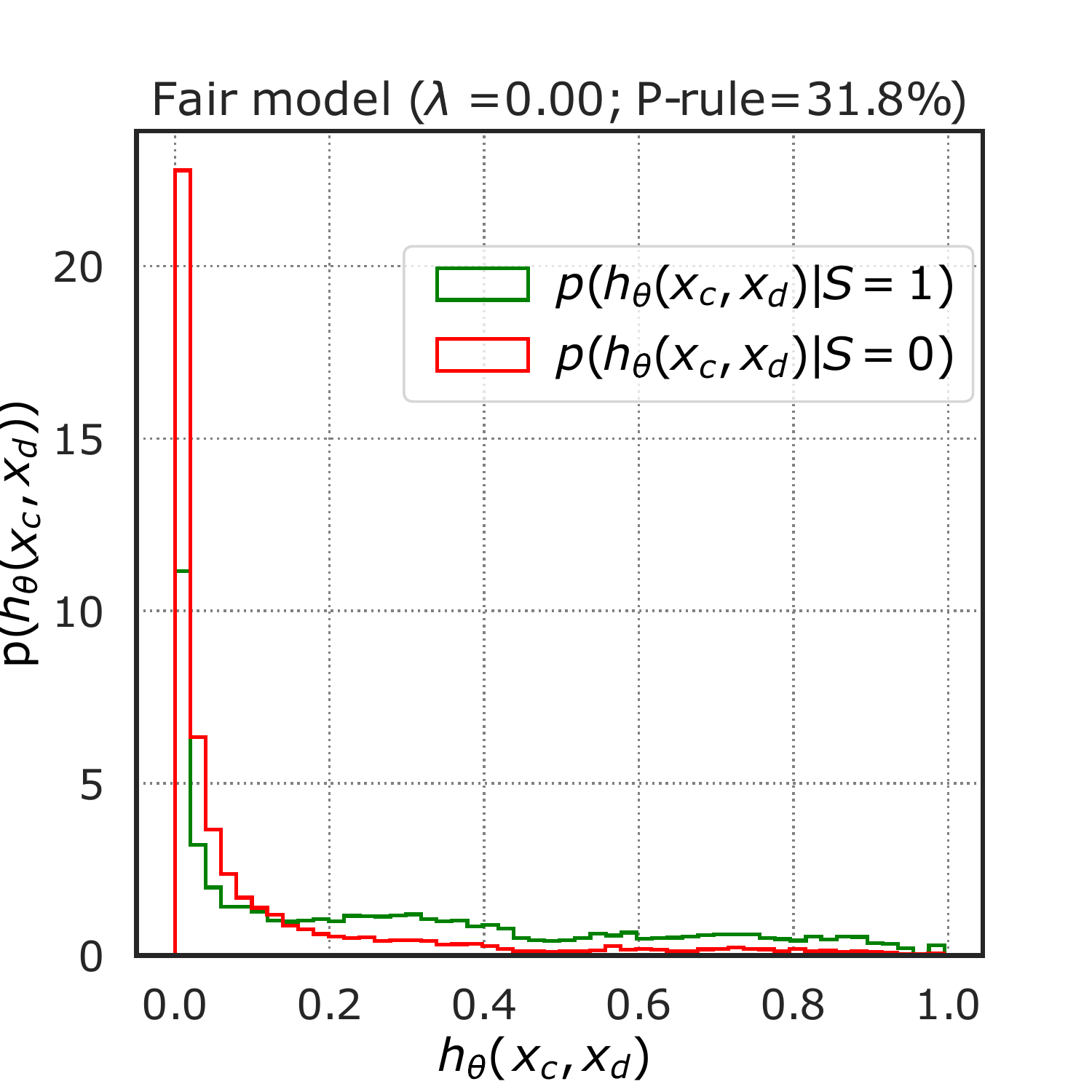}
   \includegraphics[scale=0.289,valign=t]{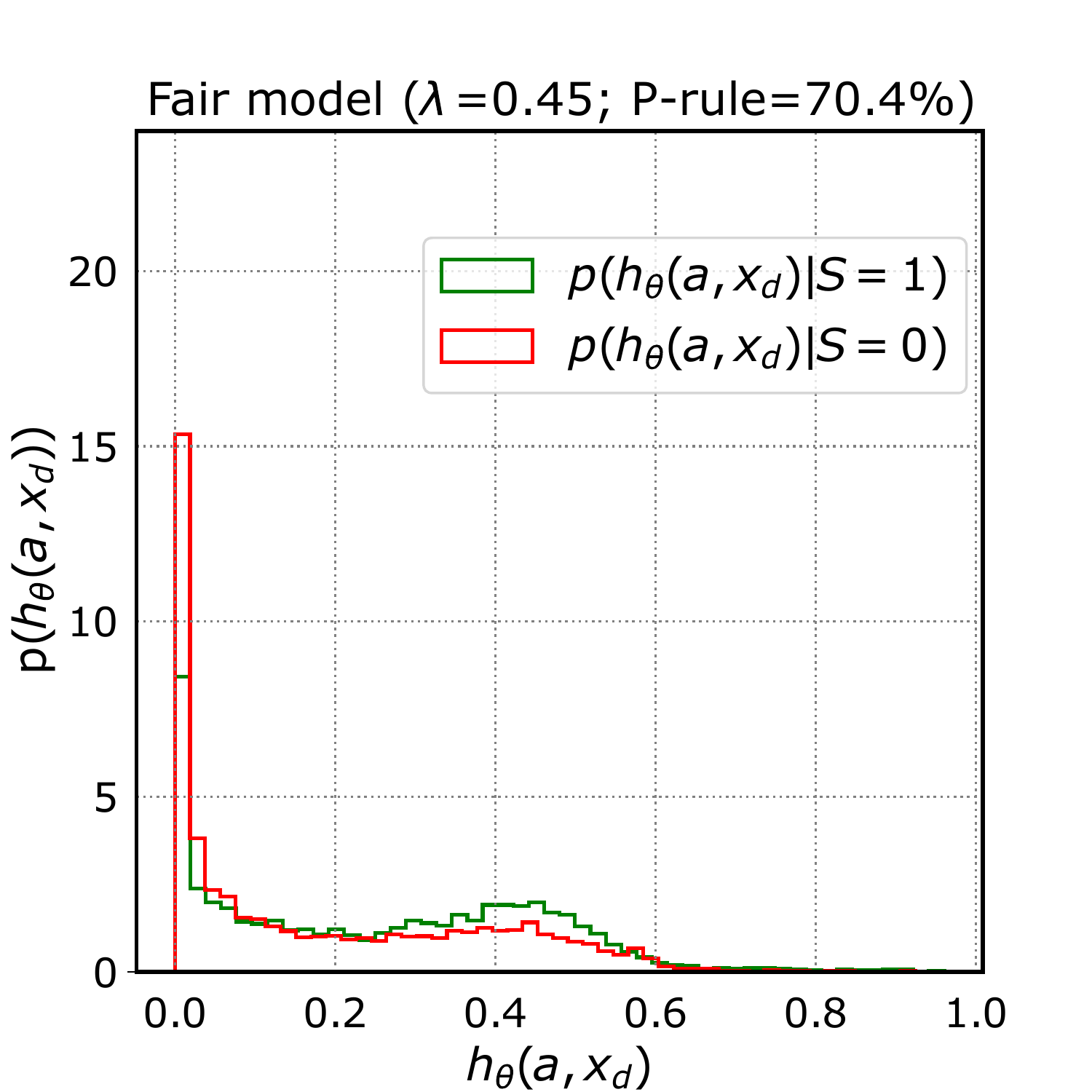}
   \includegraphics[scale=0.289,valign=t]{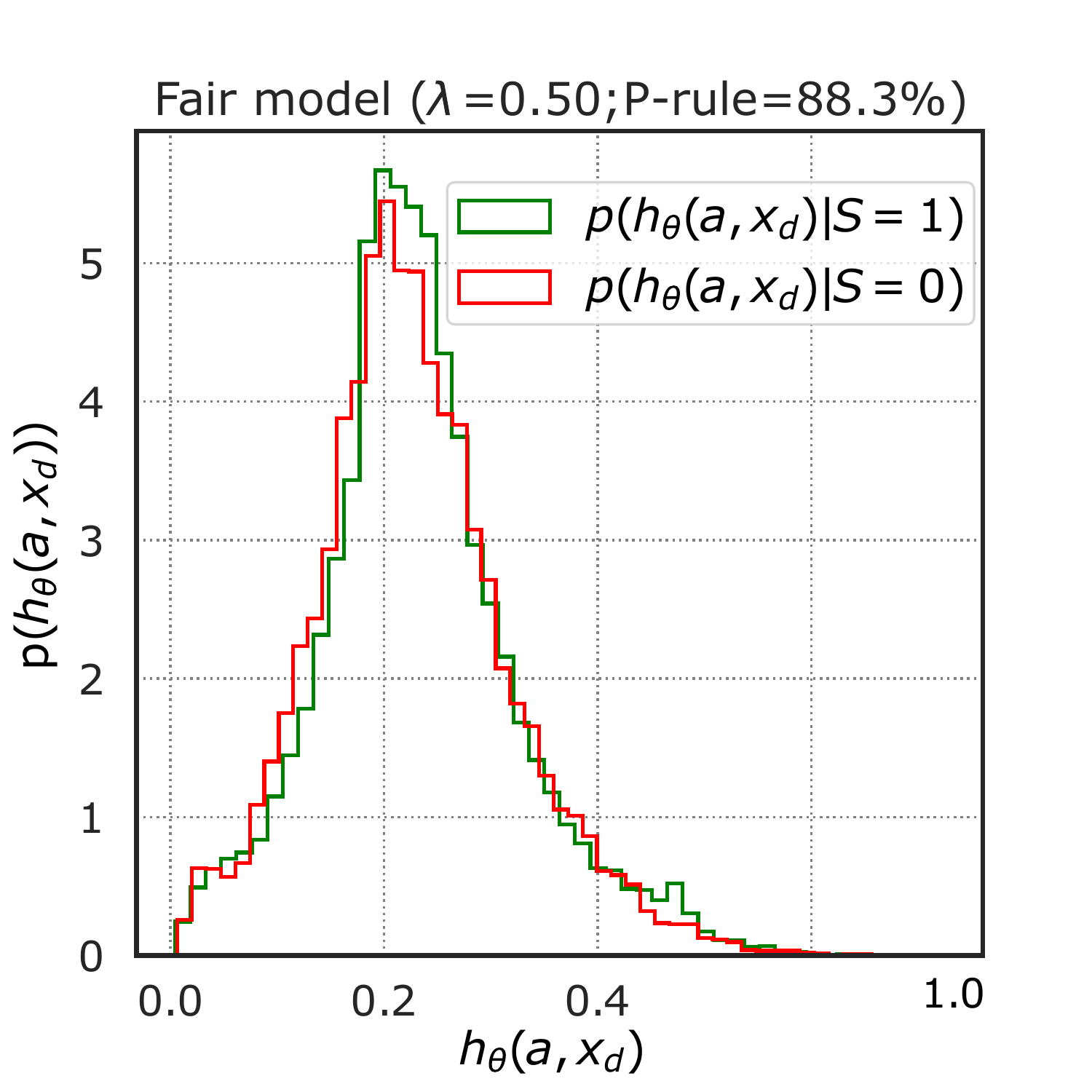}
  \caption{Distributions of the predicted probabilities given the real sensitive attribute S (Adult UCI data set)}
  \label{fig:distribution_of_predictions_by_sensitive}
\end{figure*}

\begin{figure}[H]
  \centering
  \subfloat[No mitigation ($\lambda_{DP}=0$)]{\includegraphics[scale=0.2845]{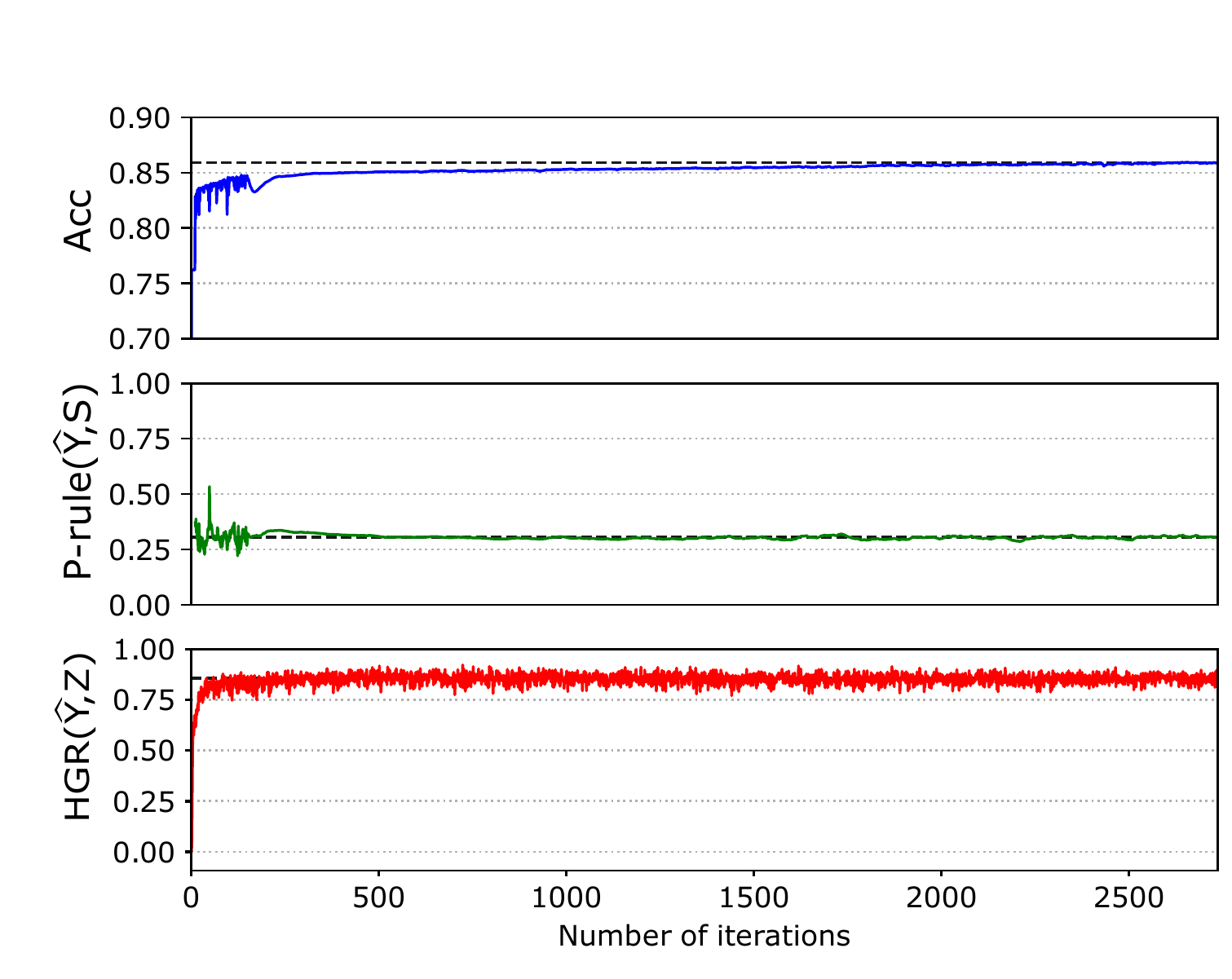}}
   \subfloat[Mitigation with $\lambda_{DP}=0.5$]{\includegraphics[scale=0.2845]{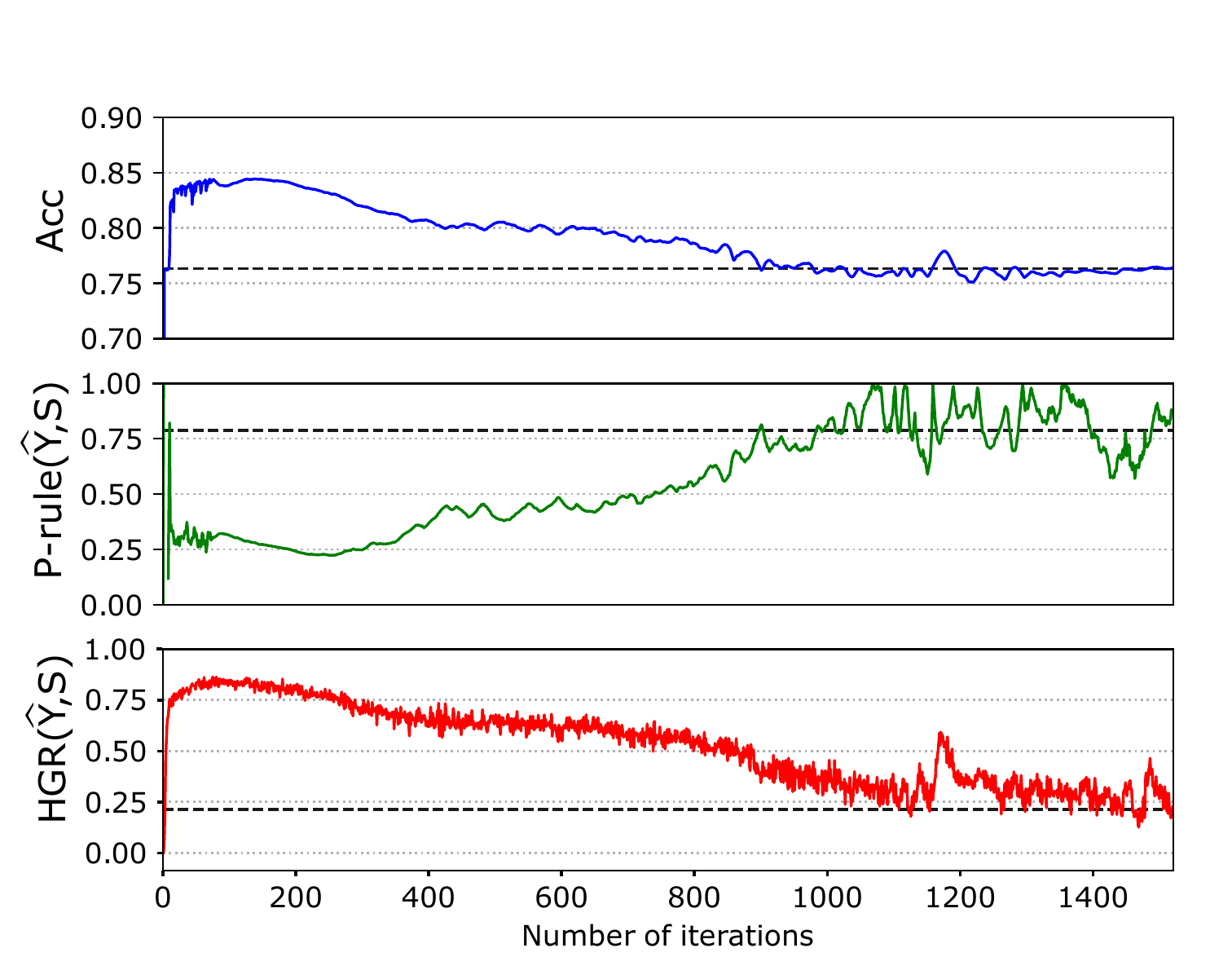}}
    \caption{Dynamics of adversarial training}
    \label{fig:trainingcurves2}
\end{figure}

The dynamics of adversarial training for demographic parity performed for Adult UCI is illustrated in Figure~\ref{fig:trainingcurves2} with an unfair model ($\lambda_{DP}=0$) and a fair model with an hyperparameters $\lambda_{DP}=0.5$ 
(results for other values are presented in appendix). 
We represent the accuracy of the model (top), the P-rule metric between the prediction and the real sensitive $S$ (middle), and the HGR between the prediction and the latent space $Z$ (bottom). 
We observe for the unfair model (leftmost graph) that the convergence is stable and achieve a P-rule to $29.5\%$. 
As desired, by incresing the hyperparameter $\lambda_{DP}$ the penalization loss decreases (measured with the $HGR$), it allows to increase the fairness metric P-rule to $83.1\%$ with a light drops of accuracy.

In Figure~\ref{fig:distribution_of_predictions_by_sensitive} we plot the distribution of outcomes depending on the sensitive value $S$ for 3 models for the demographic parity criterion and three different values of fairness weight $\lambda_{DP}$: $0$, $0.45$ and $0.50$.
For the leftmost graph (i.e. $\lambda_{DP}=0$) the model looks very unfair, since the distribution importantly differs between sensitive groups. As desired, we observe that with increased $\lambda_{DP}$ values, the distributions are more aligned.

 \paragraph{Comparison against the State-of-the-Art}

\begin{figure}[t]
\subfloat[Adult UCI Data set]{\includegraphics[scale=0.25]{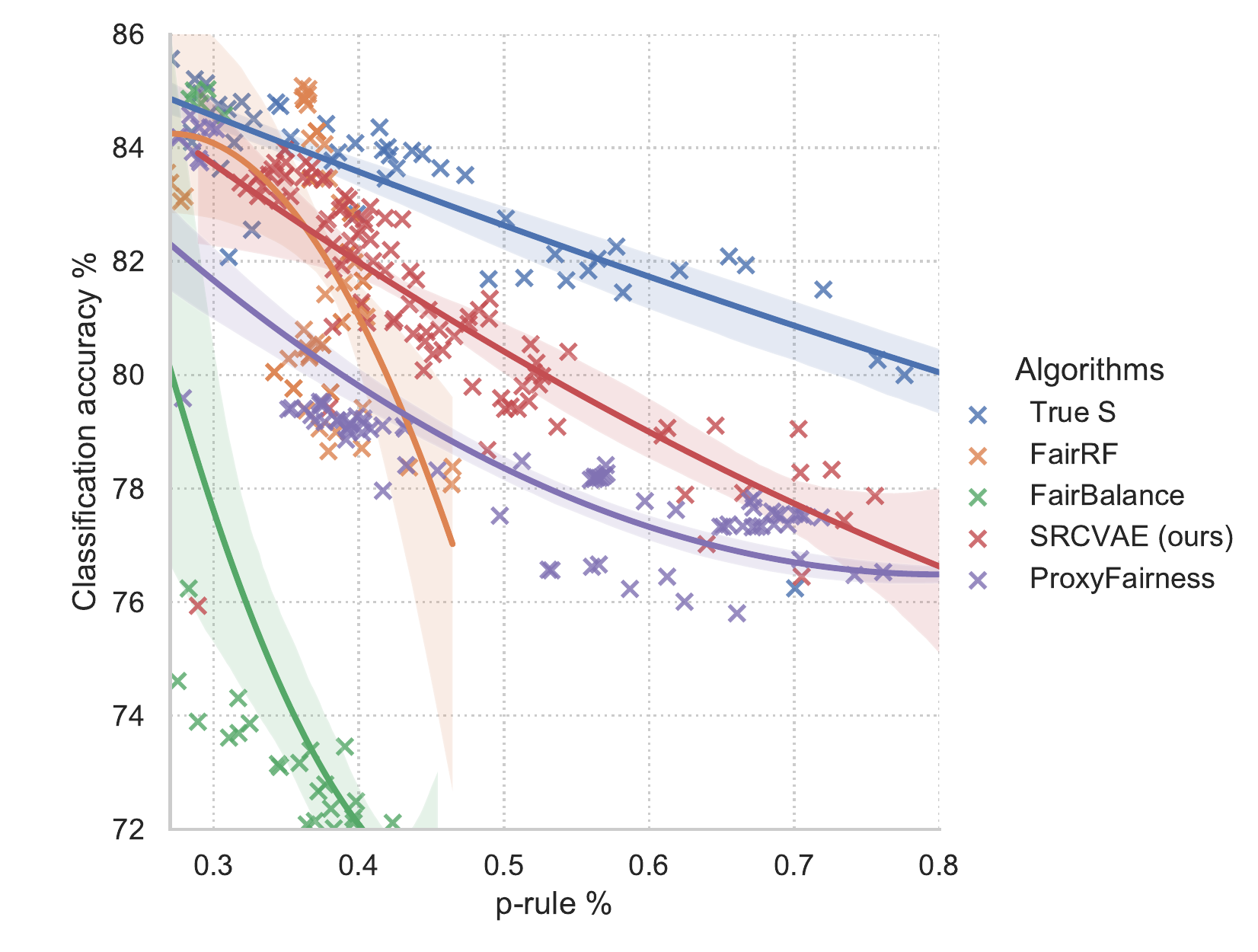}
}
\subfloat[Default Data set]{\includegraphics[scale=0.25]{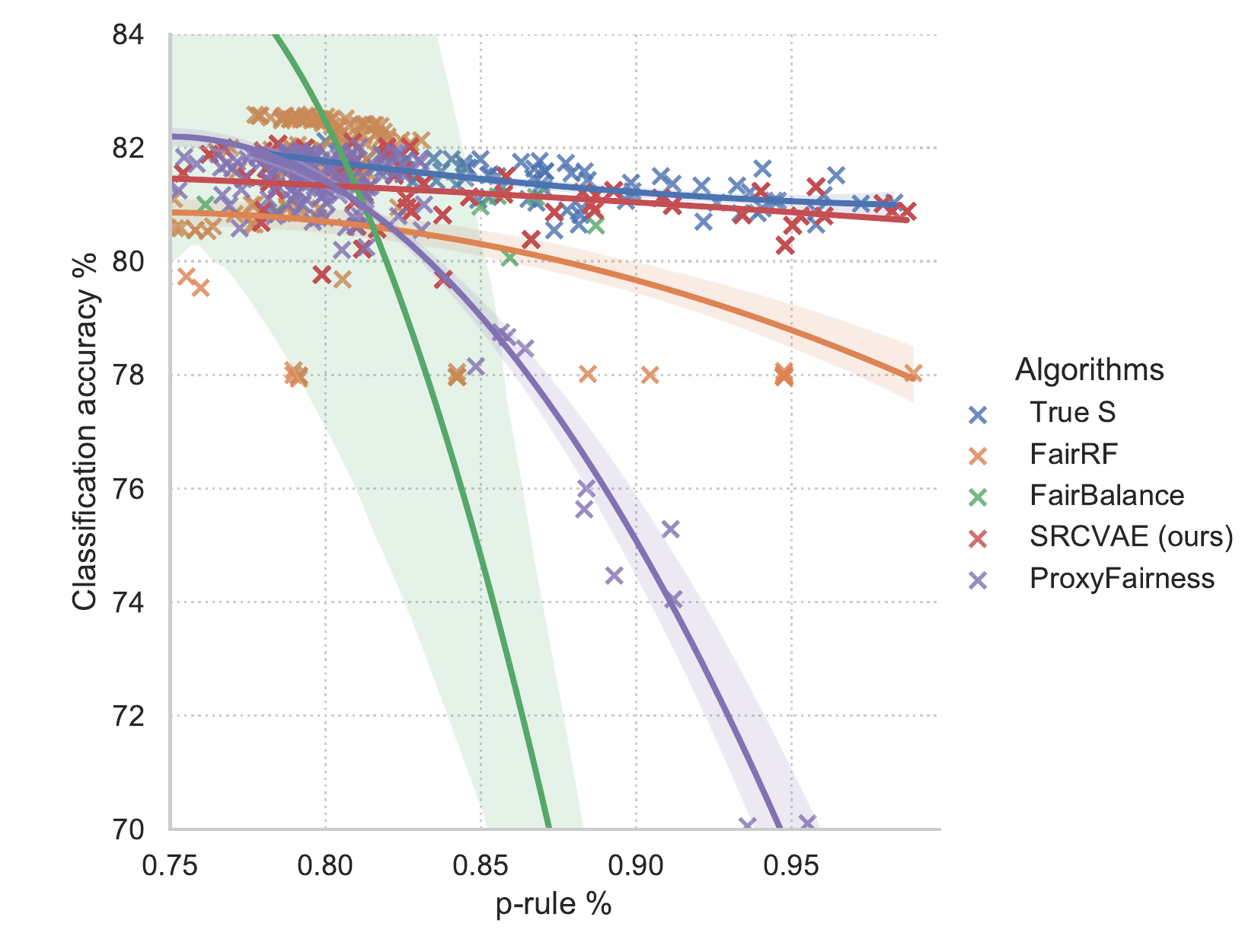}}
\caption{Demographic Parity task} 
\label{fig:demparitytask}
\end{figure}
\noindent
\begin{figure}[t]
\subfloat[Adult UCI Data set]{
\includegraphics[scale=0.255]{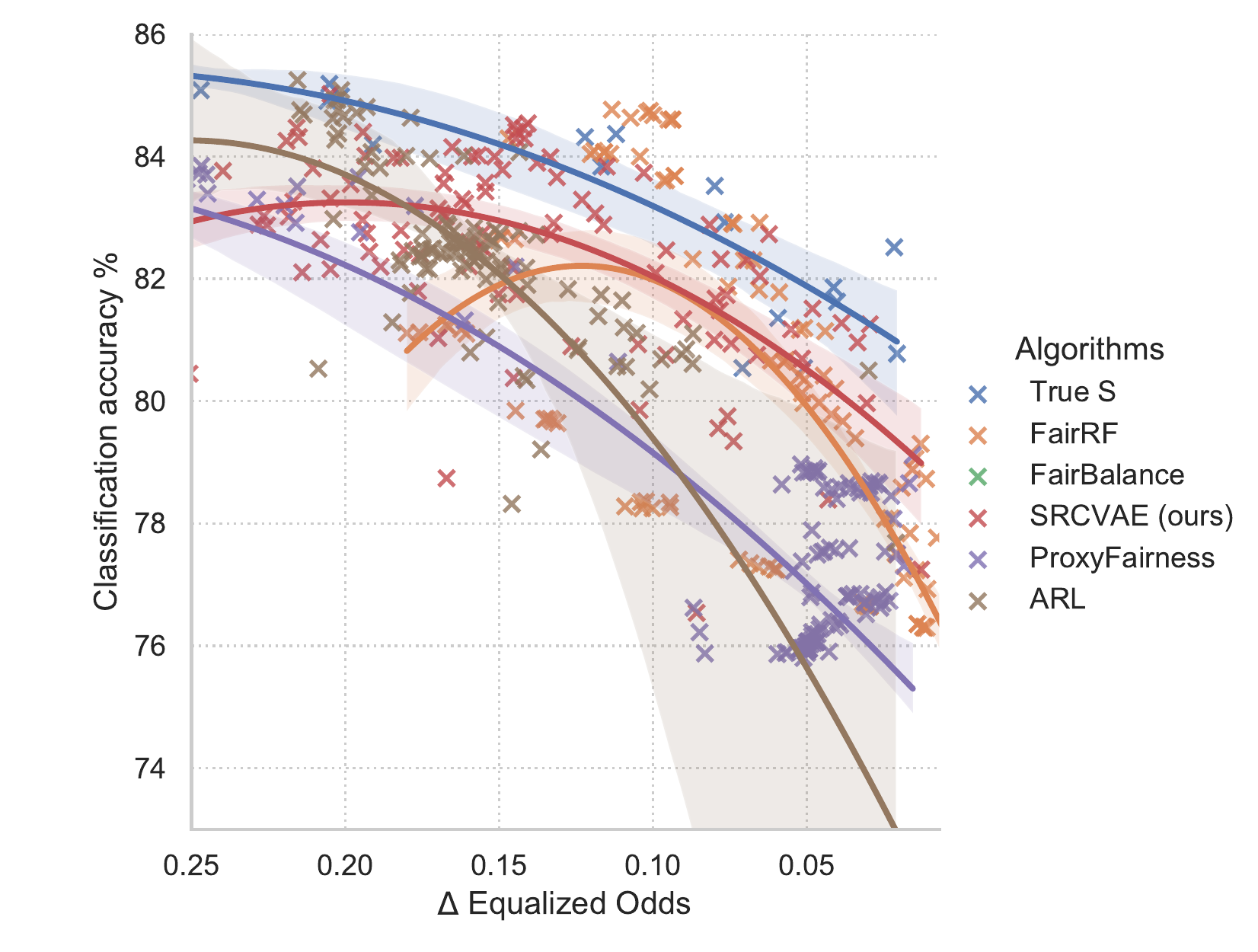} 
} 
\subfloat[Default Data set ]{\includegraphics[scale=0.255]{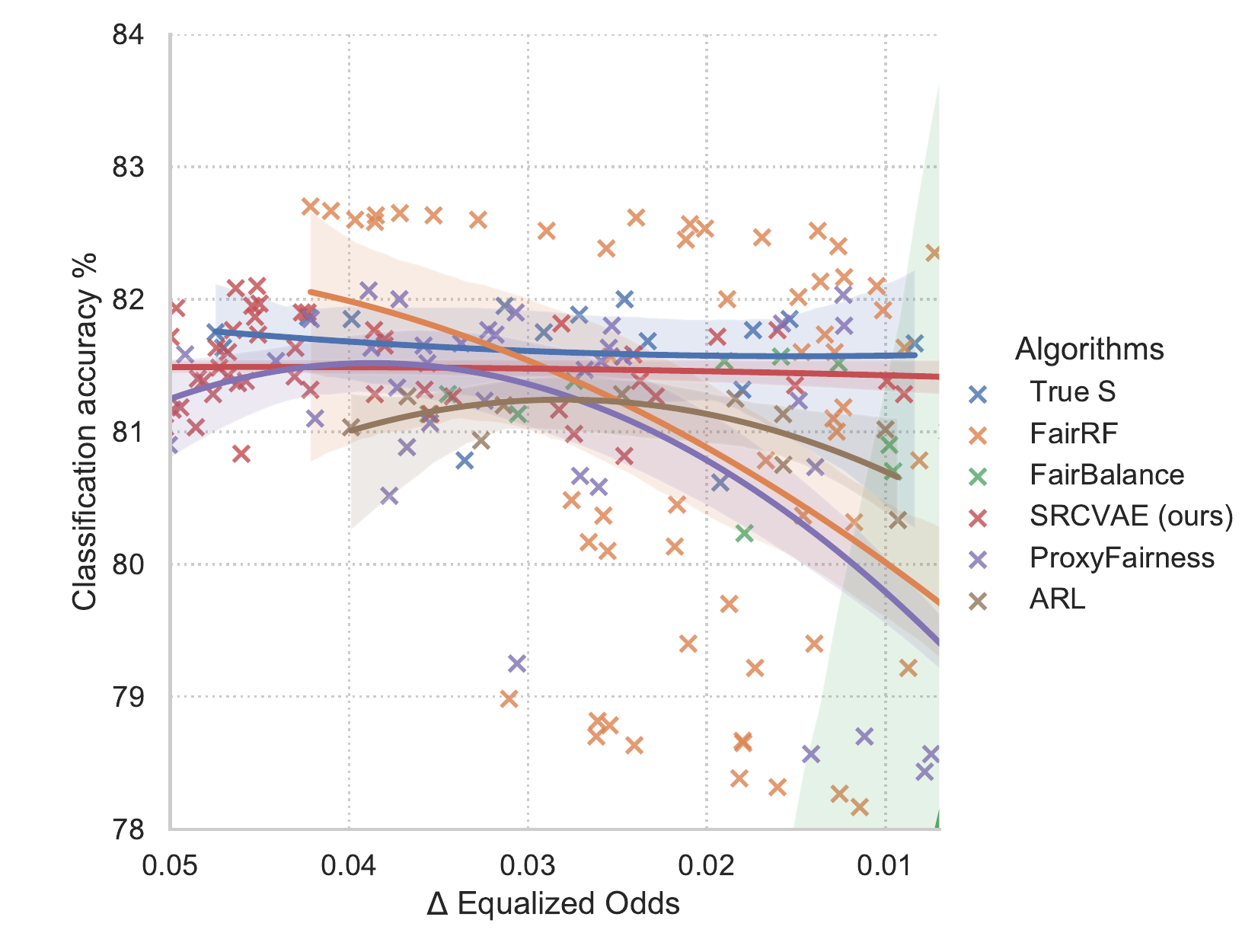} }
\caption{Equalized odds task 
}
\label{fig:equalizedoddstask}
\end{figure}




 
\noindent
For the two datasets, we experiment different models where, for each, we repeat five runs by randomly sampling two subsets, 80\% for the training set and 20\% for the test set. Because different optimization objectives result in different algorithms, we run separate experiments for the two fairness objectives of our interest. 
As an optimal baseline to be reached, we consider the approach from  \cite{adel2019one} using observations of the 
sensitive $S$ during training, which we denote as 
\emph{True S}. 
We also compare various approaches specifically designed to be trained in the absence of the 
sensitive information during the training: 
\emph{FairRF}~\citep{zhao2021you}
,
\emph{FairBalance}~\citep{yan2020fair}
, 
\emph{ProxyFairness}~\citep{gupta2018proxy} 
and \emph{ARL}~\citep{lahoti2020fairness}, where the latter is only compared for the equalized odds task (i.e. discussion in \cite{zhao2021you}).
We plot the performance of these different approaches by displaying the Accuracy against the P-rule for Demographic Parity (Figure~\ref{fig:demparitytask}) and the Disparate Mistreatment (DM) (corresponding to the sum of $\Delta_{FPR}$ and $\Delta_{FNR}$) for Equalized Odds (Figure~\ref{fig:equalizedoddstask}).
We clearly observe for all algorithms that the Accuracy, or predictive performance, decreases when fairness increases. As expected, the baseline \emph{True S} achieves the best performance for all the scenarios with the highest accuracy and fairness. 
We note that, for all levels of fairness (controlled by the mitigation weight in every approach), 
our method outperforms state-of-the-art algorithms for both fairness tasks (except some points for very low levels of fairness, at the left of the curves).   
We attribute this to the ability of SRCVAE for extracting a useful sensitive proxy, while 
the approaches \emph{FairRF} and \emph{ProxyFairness} seem to greatly suffer from only considering features present in the data for mitigating fairness. 
The approach \emph{FairBalance}, which pre-processed the data with clustering, seems inefficient and degrades the predictive performance too significantly.

\paragraph{Impact of proxy dimension}
 
\begin{figure}
\centering
\subfloat[Adult UCI Data set]{\includegraphics[scale=0.28]{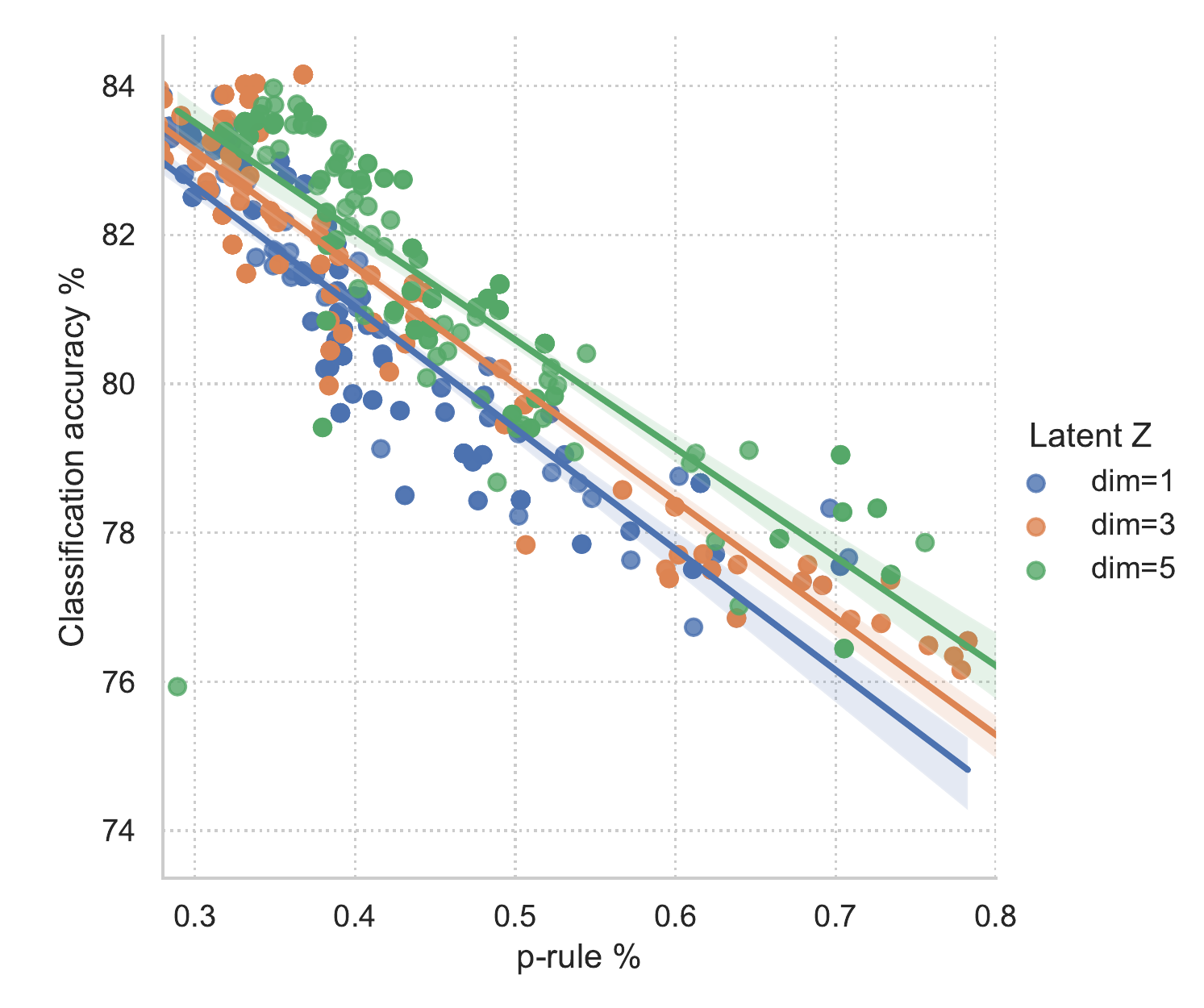}
}
\subfloat[Default Data set]{\includegraphics[scale=0.28]{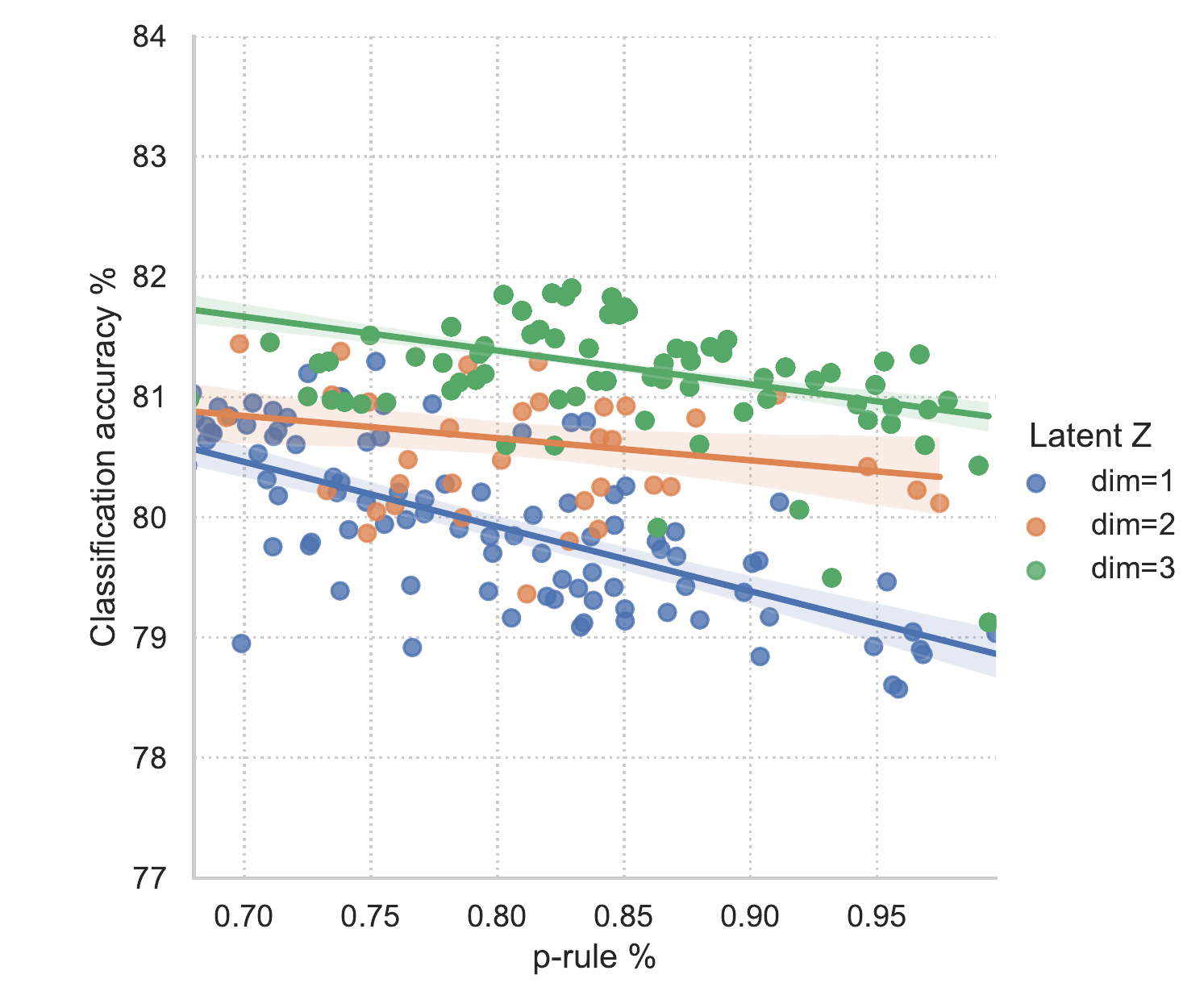}}
\caption{
Dimension of $z$ on demographic parity task} 
\label{fig:dimensions}
\end{figure}
In figure \ref{fig:dimensions}, we performs additional experiment about the sensitive proxy. We observe that using a single dimension of $z$ is less efficient than larger dimension, for the two datasets.
Increasing the dimensions to $5$ and $3$ for Adult UCI and Default respectively allows to obtain better results in terms of accuracy and this for all levels of P-rule. We claim that, as already observed in another context in \cite{grari2020learning}, mitigating biases in larger spaces allow better generalisation abilities at test time, which supports the choice of considering a multivariate sensitive proxy $z$, rather than directly acting on a reconstruction of $s$ as a univariate variable.


\section*{Conclusion and Future Work}
We present a new way for mitigating undesired bias without the availability of the sensitive demographic information in training.
By assuming a causal graph, we rely on a new variational auto-encoding based framework named SRCVAE for generating a latent representation which is expected to contain the most sensitive information as possible.
These inferred 
proxies serve, in a second phase, for bias mitigation in an adversarial fairness training of a prediction model.
Compared with other state-of-the-art algorithms, our method proves to be more efficient in terms of accuracy for 
similar levels of fairness. 
For further investigation, we are interested in extending this work to settings 
where the actual sensitive can be continuous (e.g. age or weight attribute) and/or multivariate. 

\newpage
\medskip
\small
\bibliography{MyCollection}
\end{document}


\maketitle

\begin{abstract}


This is the supplementary material of paper Fairness without the sensitive attribute via Causal Variational Autoencoder

\end{abstract}

\begin{theorem}
For two nonempty index set $S$ and $Z$ such that $S \subset Z$ and $\hat{Y}$ the output prediction of a predictor model, we have :\\ 
\begin{align}
HGR(\hat{Y},Z) \ge HGR(\hat{Y},S)
\end{align}
\end{theorem}

\begin{proof}
Let's assume that the set $Z^{c}$ represent all the elements of Z private of S:  $Z^{c}=Z \backslash S$,
Following the definition of the Hirschfeld-Gebelein-Renyi Maximum Correlation Coefficient, we have: 
\begin{align}
HGR(\hat{Y},Z)&= \sup_{\substack{ f:\mathcal{U}\rightarrow \mathbb{R},g:\mathcal{V}\rightarrow \mathbb{R}}} \rho(f(\hat{Y}), g(Z)) 
\end{align}

By the Cauchy inequality and by setting $f(Z) = E[g(\hat{Y})|Z]$, we
can show the equivalence characterization of the HGR:
\begin{align}
&HGR(\hat{Y},Z) = \sup_{\substack{g}} \sqrt{\frac{var(E(g(\hat{Y})|Z)])}{var(g(\hat{Y}))}}  \\ &= \sup_{\substack{g}} \sqrt{\frac{E(E[g(\hat{Y})|S,Z^{c}]^2)+E(g(\hat{Y})|S,Z^{c})^2 }{var(g(\hat{Y}))}} 
  \\
  &= \sup_{\substack{g}} \sqrt{\frac{E(E(E[g(\hat{Y})|S,Z^{c}]^2|S))+E(g(\hat{Y})|S,Z^{c})^2) }{var(g(\hat{Y}))}} 
 \\ & \ge \sup_{\substack{g}} \sqrt{\frac{E(E[g(\hat{Y})|S]^2)+E(g(\hat{Y})|S,Z^{c})^2 }{var(g(\hat{Y}))}} = HGR(\hat{Y},S) 
\end{align}
We use in (10) the Jensen inequality for conditional expectation.
\end{proof}

\section{Evidence Lower Bound for the left causal graph}
\begin{figure}
  \centering
  \includegraphics[scale=0.407]{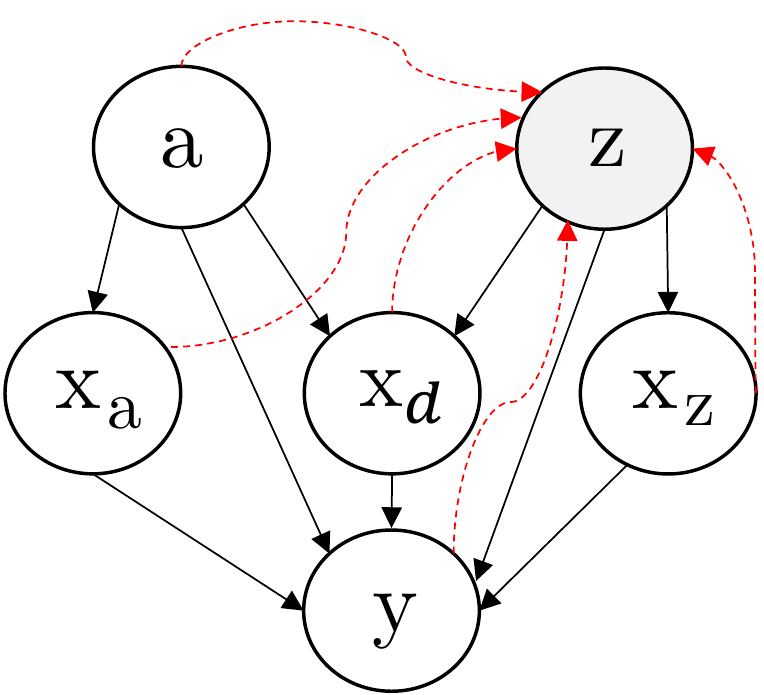}
  \caption{Causal graphs of the SRCVAE: The following causal graph represents the more general representation where $x$ is mapped into four components $x_a$, $x_d$, $x_z$ and $a$;
  }
  \label{fig:causal_graph}
\end{figure}

By following the causal graph depicted in Figure~\ref{fig:causal_graph}, the decoder distribution, $p_{\theta}(a, x_{a}, x_{d}, x_{z}, y|z)$, can be factorized as below:
\begin{align}
p_{\theta}(a, x_{a}, x_{d}, x_{z}, y|z) = p(a)p(x_{a}|a)p_{\theta}(x_{d}|a,z)p_{\theta}(x_{z}|z)p_{\theta}(y|a,x_{a},x_{d},x_{z},z) \nonumber
\end{align}
Given an approximate posterior $q_{\phi}(z|a,x_{a},x_{d},x_{z},y)$, we
obtain the variational lower bound as Eq.~
\ref{equ:elbo}
{\footnotesize
\begin{align}
log(p_{\theta}(a, x_{a}, x_{d}, x_{z}, y)) \geq & \mathbb{E}_{\substack{(a,x_{a},x_{d},x_{z},y)\sim\mathcal{D}, \\ z \sim q_{\phi}(z|a,x_{a},x_{d},x_{z},y)}}[\log p_{\theta}(x_{d}|a,z)\log p_{\theta}(y|a,x_{a},x_{d},x_{z},z)\log p_{\theta}(x_{z}|z) \nonumber \\
& 
- D_{KL}(q_{\phi}(z|a,x_{a},x_{d},x_{z},y)||p(z))\big] \nonumber \\
&=: -\mathcal{L}_{ELBO}
\label{equ:elbo}
\end{align}
}
where $D_{KL}$ denotes the Kullback-Leibler divergence of the posterior $q_{\phi}(z|a,x_{a},x_{d},x_{z},y)$ from a prior $p(z)$, typically a standard Gaussian distribution ${\cal N}(0,I)$. The posterior  $q_{\phi}(z|a,x_{a},x_{d},x_{z},y,z)$ is represented by a deep neural network with parameters $\phi$, which typically outputs the mean $\mu_\phi$ and the variance $\sigma_\phi$ of a diagonal Gaussian distribution ${\cal N}(\mu_\phi,\sigma_\phi I)$. The likelihood term factorizes as  
$p_{\theta}(x_{d},x_{z},y|a,x_{a},x_{d},x_{z},z)=  p_{\theta}(x_{d}|a,z)p_{\theta}(x_z|z)p_{\theta}(y|a,x_a,x_d,x_z,z)$, are defined as neural networks 
with parameters $\theta$. 
The maximization of this marginal log-likelihood
is realized by the minimization of the negative lower bound, mentioned as $\mathcal{L}_{ELBO}$.
Since attracted by a standard prior, the posterior is supposed to remove  probability mass for any features of  
$z$ that are not involved in the reconstruction of $x_d$, $x_z$ and $y$. Since $a$ is given together with $z$ as input of the likelihoods, all the information from $a$ should be removed from the posterior  distribution of $z$.
In practice, we use the neural network layers to infer the parameters of a Gaussian distribution over the joint space of $z$. To obtain the posterior distribution of $q_\phi$, $p(z)$ is the prior distributions following the Gaussian distribution, and we utilize the reparametrization, accordingly. Notice that the complementary set $a$ 
is not involved in any specific reconstruction. %

In addition, we employ in this paper a variant of the ELBO optimization as done in  \cite{pfohl2019counterfactual}, where the $D_{KL}(q_{\phi}(z|a,x_a,x_d,x_z,y)||p(z))$ term is replaced by a MMD term $\mathcal{L}_{MMD}(q_{\phi}(z)||p(z))$ between the aggregated posterior $q_{\phi}(z)$ and the prior. This has been shown more powerful than the classical $D_{KL}$  
for ELBO optimization in \cite{zhao2017infovae}, as the latter can reveal as 
too restrictive (uninformative latent code problem) \cite{chen2016variational,bowman2015generating,sonderby2016ladder} and can also tend to overfit the data (Variance Over-estimation in Feature Space).


This inference must however ensure that no
dependence is created between $a$ and $z$ (no arrow from $a$ to $z$ in the graph from \ref{fig:causal_graph}), unless preventing the generation of proper sensitive proxy which is not linked to the complementary.
However, 
by optimizing this ELBO optimization, some dependence can still be observed empirically between $a$ and $z$. 
Some information from $a$ leaks in the inferred $z$. In order to ensure some minimum independence level we add a dependence penalisation term in this loss function. Leveraging the last research for mitigating the dependence with continuous multidimensional space we extend the main idea of \cite{grari2020learning} by adapting this penalization in the variational autoencoder case. Originally, this paper used an HGR estimation in a minmax game to penalize the intrinsic bias in a multi dimensional latent representation for deterministic autoencoder. They have showed that a neural HGR-based approach presents a very competitive results
in the continuous case by identifying some
optimal transformations for multidimensional features. 

Finally, the inference of our SRCVAE is optimized by a mini-max game as follows: 

\begin{eqnarray}
\begin{aligned}
\nonumber
    \argmin_{\theta,\phi}\max_{w_{f},w_{g}}&-\mathbb{E}_{\substack{(a,x_{a},x_{d},x_{z},y)\sim\mathcal{D}, \\ z \sim q_{\phi}(z|a,x_{a},x_{d},x_{z},y)}}[[\log p_{\theta}(x_{d}|a,z)\log p_{\theta}(y|a,x_{a},x_{d},x_{z},z)\log p_{\theta}(x_{z}|z)] \\
    &+\lambda_{mmd}\mathcal{L}_{MMD}(q_{\phi}(z)||p(z))  \\
    &+
    \lambda_{inf}\mathbb{E}_{\substack{(a,x_{a},x_{d},x_{z},y)\sim\mathcal{D}, \\ z \sim q_{\phi}(z|a,x_{a},x_{d},x_{z},y)}}(\widehat{f}_{w_{f}}(a)\widehat{g}_{w_{g}}(z))]
    \label{FinalInference}
\end{aligned}
\end{eqnarray}

where $\lambda_{mmd}$, $\lambda_{inf}$ are scalar hyperparameters. 
The additional MMD objective  
can be interpreted as minimizing the distance between all moments of each aggregated latent code distribution and the prior distribution. 
Note that 
the use of $y$ as input for our generic inference scheme $q(z|a,x_a,x_d,x_z,y)$ is allowed since $z$ is only used during training for learning a fair predictive model and is not used at deployment time. 

The complementary set $a$ is the only input given to the adversarial $f$ and the continuous latent space $z$ as input for the adversarial $g$. In that case, we only capture for each gradient iteration the estimated HGR between the complementary set and the generated proxy latent space. The algorithm takes as input a training set from which it samples batches of size $b$ at each iteration. At each iteration it first standardize the output scores of networks $f_{\theta_f}$ and $g_{\theta_g}$ to ensure 0 mean and a variance of 1 on the batch. Then it computes the objective function to maximize to estimate the HGR score and the global variational reconstruction objective. At the end of each iteration, the algorithm updates the parameters of the encoders parameters $\theta$ as well as the encoder parameters $\phi$ by one step of gradient descent. Concerning the HGR adversary, the backpropagation of the parameters $\theta_f$ and $\theta_g$ is performed by multiple steps of gradient ascent. This allows us to optimize a more accurate estimation of the HGR at each step, leading to a greatly more stable learning process. 


\section{Experiments}
For our experiments we use 2 different popular datasets often used in fair classification:   
\begin{itemize}
\item Adult: The Adult UCI income data set~\cite{Dua:2019} contains 14 demographic attributes of approximately 45,000 individuals together with class labels which state if their income is higher than \$50,000 or not. As sensitive attribute we use gender encoded as a binary attribute, male or female. As complementary set we use Race, Age and Native country and the descendant set from all the others features (education, marital status, occupation..).

\item Default: The Default data set~\cite{Yeh:2009:CDM:1464526.1465163} contains 23 features about 30,000 Taiwanese credit card users with class labels which state whether an individual will default on payments. As sensitive attribute we use gender encoded as a binary attribute, male or female. The complementary set is composed of the continuous feature age and the descendant set from all others features (household size, education level, marital status).

\end{itemize}

\begin{figure}
   \centering
    \subfloat[$\lambda=0.00$ ;  $P-rule=29.5\%$]{\includegraphics[scale=0.5]{pictures/Trainingcurves_lam0P-rule=29.0 (8).pdf}}
    \subfloat[$\lambda=0.24$ ;  $P-rule=32.1\%$]{\includegraphics[scale=0.5]{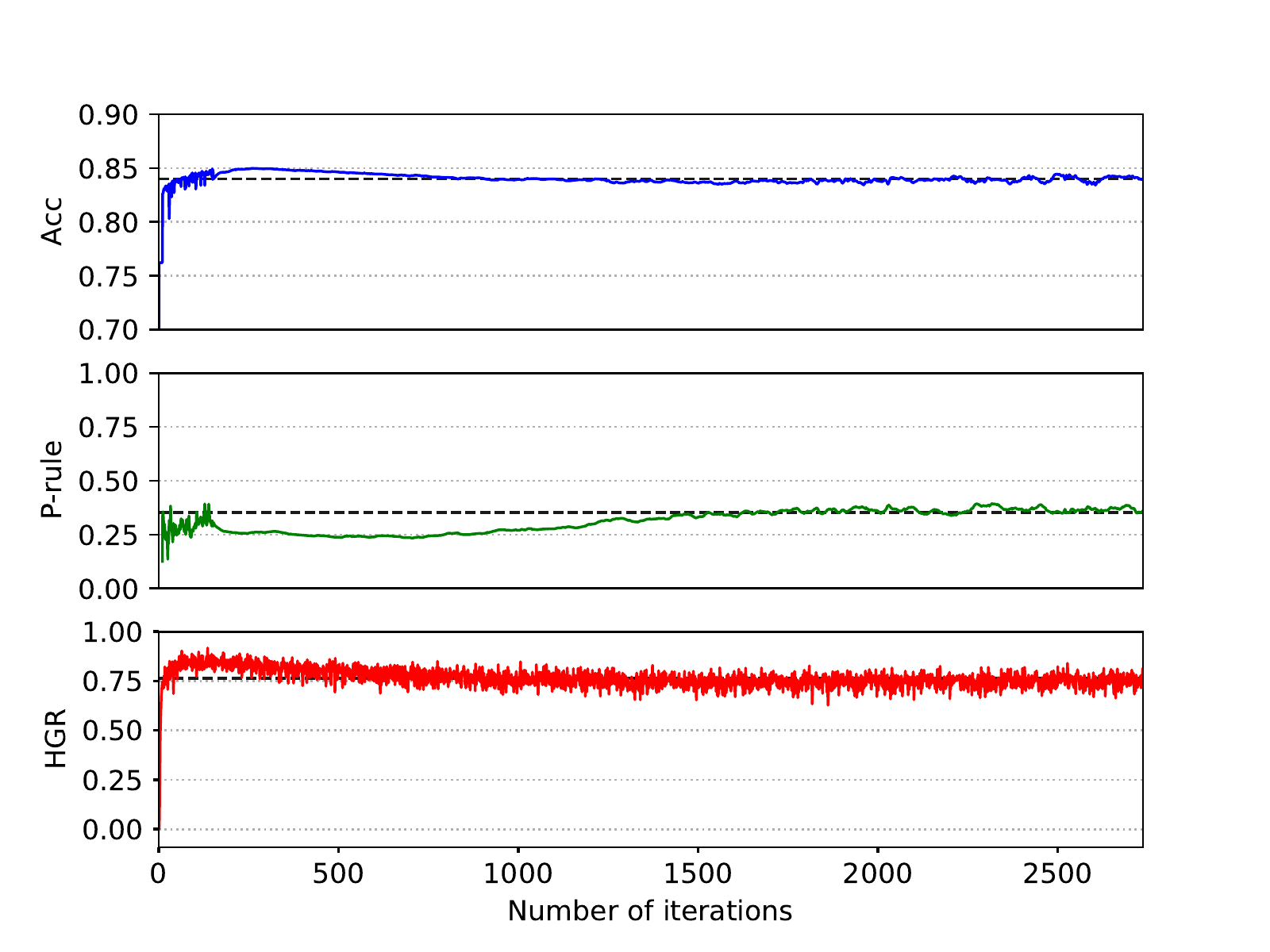}} \\
    \subfloat[$\lambda=0.35$ ;  $P-rule=49.5\%$]{\includegraphics[scale=0.5]{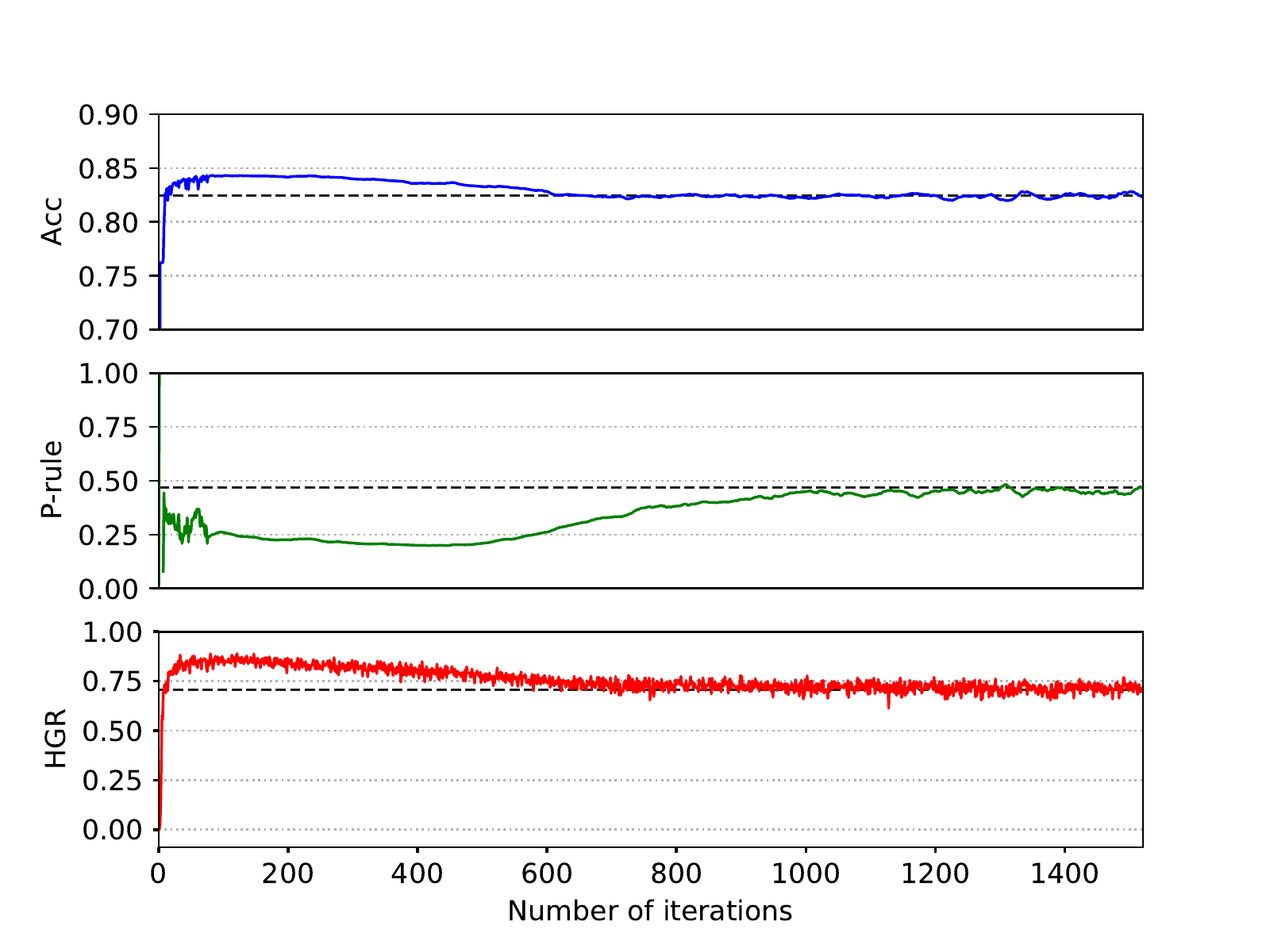}}
    \subfloat[$\lambda=0.45$ ;  $P-rule=58.9\%$]{\includegraphics[scale=0.5]{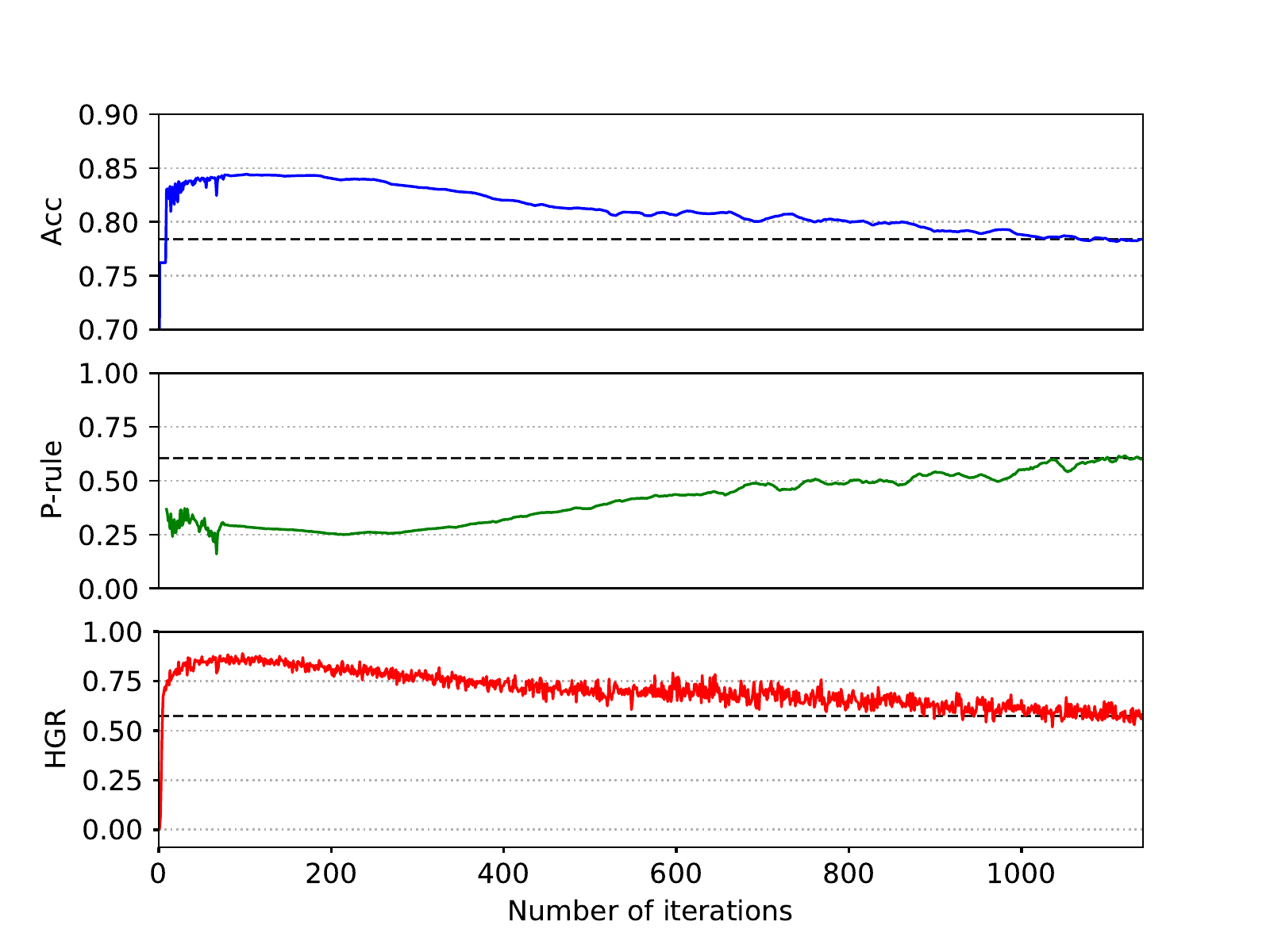}}\\
    \subfloat[$\lambda=0.48$ ;  $P-rule=65.4\%$]{\includegraphics[scale=0.5]{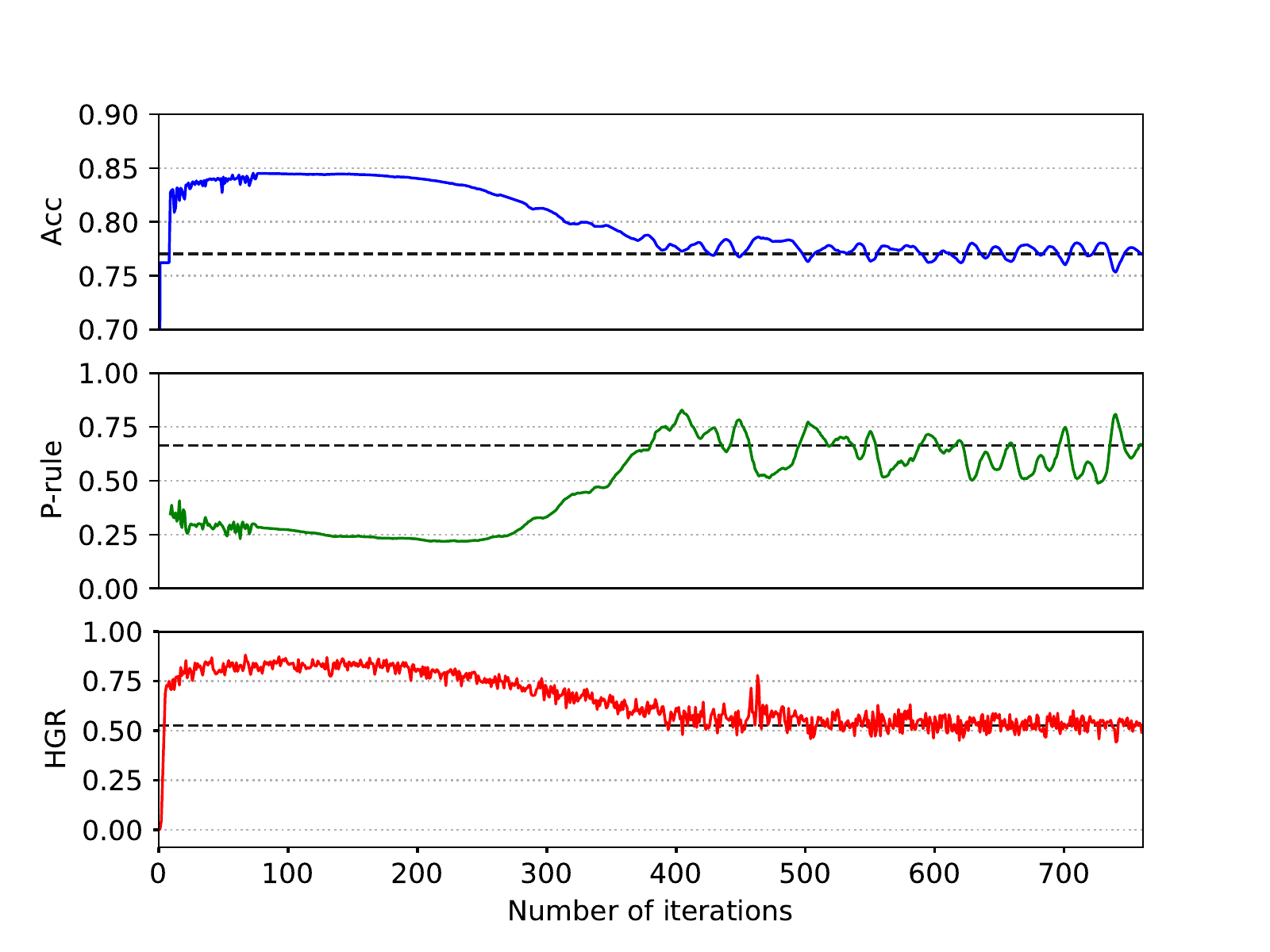}}
    \subfloat[$\lambda=0.5$ ;  $P-rule=82.9\%$]{\includegraphics[scale=0.5]{pictures/Trainingcurves_lam0.5P-rule=97.9 (4).pdf}}

  \caption{Training curves with different hyperparameters $\lambda_{DP}$}
     \label{fig:different_hyp}
\end{figure}

We present in Figure~\ref{fig:different_hyp} the dynamics of the adversarial training with different hyperparameters $\lambda_{DP}$,  optimized  for  demographic  parity. The choice of this value depends on the main objective, resulting in  a  trade-off between accuracy and fairness. 
We observe that higher values of $\lambda_{DP}$ produce fairer predictions, the P-rule (assessed with the real sensitive) increase, while $\lambda_{DP}$ near 0 allows to only focus on optimizing the classifier predictor.

\newpage
\bibliography{MyCollection}
